\documentclass[12pt]{article}
\usepackage[T1]{fontenc} % modern encoding

\usepackage{mathptmx}  % uses Times for text and mathematics
\usepackage[scaled]{helvet} %  helvetica for sanserif, scaled 95% by default
\usepackage{courier}  % courier for typewriter font, pretty ugly but available
\usepackage{cite}  % improves format of numerical citations, such as [1-3]
\usepackage[top=1.8cm, bottom=1.8cm,left=4.0cm,right=1.5cm]{geometry}
\usepackage[onehalfspacing]{setspace}  % gives one-and-a-half line spacing
\usepackage{cite}
\usepackage[dvipsnames]{xcolor}
\usepackage{graphicx} % Required for inserting images
\usepackage{hyperref}
\usepackage{booktabs}
\usepackage{comment}
\usepackage{float}
\usepackage{tabularx}
\usepackage{makecell}
\usepackage{pifont}% http://ctan.org/pkg/pifont
\usepackage{bm} % in the preamble
\usepackage{amsmath,amssymb,amsfonts}

\usepackage{subcaption}
\usepackage{textcomp}
\definecolor{fuchsia}{rgb}{155, 0, 255}  % pure magenta
\definecolor{1}{RGB}{255,0,136}
\definecolor{2}{RGB}{255,255,80}
\definecolor{3}{RGB}{255,119,0}
\definecolor{4}{RGB}{0,255,221}
\definecolor{5}{RGB}{187,0,255}
\definecolor{6}{RGB}{0,170,255}
\definecolor{7}{RGB}{0,255,136}
\definecolor{8}{RGB}{102,0,255}
\definecolor{9}{RGB}{0,102,255}
\definecolor{10}{RGB}{0,255,0}
\definecolor{11}{RGB}{255,0,0}
\definecolor{12}{RGB}{0,255,53}
\definecolor{13}{RGB}{61,255,80}
\definecolor{14}{RGB}{200,240,55}
\definecolor{15}{RGB}{240,121,50}

\def\BibTeX{{\rm B\kern-.05em{\sc i\kern-.025em b}\kern-.08em
    T\kern-.1667em\lower.7ex\hbox{E}\kern-.125emX}}

\usepackage{tikz}
\usepackage{graphicx}
\usepackage{subcaption}
\usepackage{caption}

\usepackage{algorithm2e}
\usepackage{amssymb}
\usepackage{amsmath}
\usepackage{booktabs}
\usepackage{comment}
\usepackage{float}
\usepackage{tabularx}
\usepackage{makecell}
\usepackage{pifont}% http://ctan.org/pkg/pifont
\usepackage{url} %links
\usepackage{multirow}
\usepackage{pgfplotstable}

\definecolor{16}{RGB}{240,128,128}
\definecolor{17}{RGB}{70,130,180}

\begin{document}

%An In-depth Analysis of Federated Quadruplet Learning with Stochastic Client Selection and Feature Embedding Stability

\title{Enhancing Federated Quadruplet Learning: Stochastic Client Selection and Embedding Stability Analysis}

\author{\"Ozg\"u  G\"oksu\\
School of Computing Science\\
University of Glasgow\\
{\tt\small 2718886G@student.gla.ac.uk}
\and
Nicolas Pugeault\\
School of Computing Science\\
University of Glasgow\\
{\tt\small nicolas.pugeault@glasgow.ac.uk}
}
\maketitle
\begin{abstract}

Federated Learning (FL) enables decentralised model training across distributed clients without requiring data centralisation. However, the generalisation performance of the global model is usually degraded by data heterogeneity across clients, particularly under limited data availability and class imbalance. To address this challenge, we propose FedQuad, a novel method that explicitly enforces minimising intra-class representations while enabling inter-class splits across clients. By jointly minimising distances between positive pairs and maximising distances between negative pairs, the proposed approach mitigates representation misalignment introduced during model aggregation. We evaluate our method on CIFAR-10, CIFAR-100, and Tiny-ImageNet under diverse non-IID settings and varying numbers of clients, demonstrating consistent improvements over existing baselines. Additionally, we provide a comprehensive analysis of metric learning-based approaches in both centralised and federated environments, highlighting their effectiveness in alleviating representation collapse under heterogeneous data distributions.
\end{abstract}
\section{Introduction}
As neural network architectures continue to grow in depth and complexity, their training increasingly relies on large-scale and diverse datasets. While deep models achieve remarkable performance on benchmarks such as ImageNet \cite{deng2009imagenet}, and LAION \cite{schuhmann2022laion}, such data is usually inaccessible in practice (a large number of samples causes huge memory space). In many real-world scenarios, data is inherently distributed across multiple entities, including academic institutions, research laboratories, and private organisations.
\begin{figure*}[htp]
    \centering
    \tikzset{every picture/.style={line width=0.75pt}} %set default line width to 0.75pt        

\begin{tikzpicture}[x=0.75pt,y=0.75pt,yscale=-1,xscale=1]
%uncomment if require: \path (0,235); %set diagram left start at 0, and has height of 235

%Shape: Ellipse [id:dp08347307039883167] 
\draw  [color={rgb, 255:red, 189; green, 16; blue, 224 }  ,draw opacity=0.91 ][fill={rgb, 255:red, 189; green, 16; blue, 224 }  ,fill opacity=0.18 ] (392.76,56.26) .. controls (405.04,49.46) and (421.55,50.83) .. (429.65,59.33) .. controls (437.75,67.83) and (434.37,80.23) .. (422.1,87.03) .. controls (409.82,93.83) and (393.31,92.46) .. (385.2,83.96) .. controls (377.1,75.46) and (380.49,63.06) .. (392.76,56.26) -- cycle ;
%Shape: Ellipse [id:dp059918250623160585] 
\draw  [color={rgb, 255:red, 74; green, 144; blue, 226 }  ,draw opacity=0.42 ][fill={rgb, 255:red, 74; green, 144; blue, 226 }  ,fill opacity=0.19 ] (329.11,65.44) .. controls (329.11,53.87) and (340.28,44.49) .. (354.06,44.49) .. controls (367.84,44.49) and (379.01,53.87) .. (379.01,65.44) .. controls (379.01,77.01) and (367.84,86.39) .. (354.06,86.39) .. controls (340.28,86.39) and (329.11,77.01) .. (329.11,65.44) -- cycle ;
%Shape: Ellipse [id:dp17729136488574204] 
\draw  [color={rgb, 255:red, 208; green, 2; blue, 27 }  ,draw opacity=0.74 ][fill={rgb, 255:red, 208; green, 2; blue, 27 }  ,fill opacity=0.22 ] (282.82,36.91) .. controls (282.82,25.34) and (293.99,15.96) .. (307.76,15.96) .. controls (321.54,15.96) and (332.71,25.34) .. (332.71,36.91) .. controls (332.71,48.48) and (321.54,57.86) .. (307.76,57.86) .. controls (293.99,57.86) and (282.82,48.48) .. (282.82,36.91) -- cycle ;
%Shape: Ellipse [id:dp727777757267692] 
\draw  [color={rgb, 255:red, 189; green, 16; blue, 224 }  ,draw opacity=0.91 ][fill={rgb, 255:red, 189; green, 16; blue, 224 }  ,fill opacity=0.18 ] (101.91,75.96) .. controls (101.91,64.39) and (113.08,55.02) .. (126.86,55.02) .. controls (140.63,55.02) and (151.8,64.39) .. (151.8,75.96) .. controls (151.8,87.53) and (140.63,96.91) .. (126.86,96.91) .. controls (113.08,96.91) and (101.91,87.53) .. (101.91,75.96) -- cycle ;
%Shape: Ellipse [id:dp21210073649858507] 
\draw  [color={rgb, 255:red, 187; green, 131; blue, 38 }  ,draw opacity=0.28 ][fill={rgb, 255:red, 245; green, 166; blue, 35 }  ,fill opacity=0.25 ] (261.57,89.37) .. controls (261.57,77.81) and (272.74,68.43) .. (286.52,68.43) .. controls (300.3,68.43) and (311.46,77.81) .. (311.46,89.37) .. controls (311.46,100.94) and (300.3,110.32) .. (286.52,110.32) .. controls (272.74,110.32) and (261.57,100.94) .. (261.57,89.37) -- cycle ;
%Shape: Ellipse [id:dp9160833233156125] 
\draw  [color={rgb, 255:red, 187; green, 131; blue, 38 }  ,draw opacity=0.93 ][fill={rgb, 255:red, 245; green, 166; blue, 35 }  ,fill opacity=1 ] (267,92.89) .. controls (267,90.02) and (270.17,87.69) .. (274.08,87.69) .. controls (277.98,87.69) and (281.15,90.02) .. (281.15,92.89) .. controls (281.15,95.76) and (277.98,98.08) .. (274.08,98.08) .. controls (270.17,98.08) and (267,95.76) .. (267,92.89) -- cycle ;
%Shape: Ellipse [id:dp5244174322189694] 
\draw  [color={rgb, 255:red, 187; green, 131; blue, 38 }  ,draw opacity=0.93 ][fill={rgb, 255:red, 245; green, 166; blue, 35 }  ,fill opacity=0.68 ] (289.85,99.35) .. controls (289.85,96.48) and (293.01,94.16) .. (296.92,94.16) .. controls (300.83,94.16) and (303.99,96.48) .. (303.99,99.35) .. controls (303.99,102.23) and (300.83,104.55) .. (296.92,104.55) .. controls (293.01,104.55) and (289.85,102.23) .. (289.85,99.35) -- cycle ;
%Shape: Ellipse [id:dp7542241000587592] 
\draw  [color={rgb, 255:red, 187; green, 131; blue, 38 }  ,draw opacity=0.93 ][fill={rgb, 255:red, 245; green, 166; blue, 35 }  ,fill opacity=0.45 ] (279.44,84.18) .. controls (279.44,81.31) and (282.61,78.98) .. (286.52,78.98) .. controls (290.43,78.98) and (293.59,81.31) .. (293.59,84.18) .. controls (293.59,87.05) and (290.43,89.37) .. (286.52,89.37) .. controls (282.61,89.37) and (279.44,87.05) .. (279.44,84.18) -- cycle ;
%Shape: Ellipse [id:dp46027197792387176] 
\draw  [color={rgb, 255:red, 57; green, 75; blue, 100 }  ,draw opacity=1 ][fill={rgb, 255:red, 74; green, 144; blue, 226 }  ,fill opacity=0.42 ] (335.44,57.49) .. controls (335.44,54.62) and (338.61,52.29) .. (342.52,52.29) .. controls (346.42,52.29) and (349.59,54.62) .. (349.59,57.49) .. controls (349.59,60.36) and (346.42,62.69) .. (342.52,62.69) .. controls (338.61,62.69) and (335.44,60.36) .. (335.44,57.49) -- cycle ;
%Shape: Ellipse [id:dp3821223603629196] 
\draw  [color={rgb, 255:red, 57; green, 75; blue, 100 }  ,draw opacity=1 ][fill={rgb, 255:red, 74; green, 144; blue, 226 }  ,fill opacity=1 ] (338.88,75.44) .. controls (338.88,72.57) and (342.05,70.24) .. (345.95,70.24) .. controls (349.86,70.24) and (353.03,72.57) .. (353.03,75.44) .. controls (353.03,78.31) and (349.86,80.64) .. (345.95,80.64) .. controls (342.05,80.64) and (338.88,78.31) .. (338.88,75.44) -- cycle ;
%Shape: Ellipse [id:dp8809638357235923] 
\draw  [color={rgb, 255:red, 57; green, 75; blue, 100 }  ,draw opacity=1 ][fill={rgb, 255:red, 74; green, 144; blue, 226 }  ,fill opacity=0.65 ] (354.06,65.44) .. controls (354.06,62.57) and (357.23,60.24) .. (361.14,60.24) .. controls (365.04,60.24) and (368.21,62.57) .. (368.21,65.44) .. controls (368.21,68.31) and (365.04,70.64) .. (361.14,70.64) .. controls (357.23,70.64) and (354.06,68.31) .. (354.06,65.44) -- cycle ;
%Straight Lines [id:da8321652326495128] 
\draw [color={rgb, 255:red, 74; green, 74; blue, 74 }  ,draw opacity=1 ] [dash pattern={on 4.5pt off 4.5pt}]  (260.7,34.13) -- (402.66,136.74) ;
%Shape: Ellipse [id:dp06889462025764448] 
\draw  [color={rgb, 255:red, 65; green, 117; blue, 5 }  ,draw opacity=0.6 ][fill={rgb, 255:red, 65; green, 117; blue, 5 }  ,fill opacity=0.27 ] (34.37,99.9) .. controls (34.37,88.33) and (45.54,78.95) .. (59.31,78.95) .. controls (73.09,78.95) and (84.26,88.33) .. (84.26,99.9) .. controls (84.26,111.46) and (73.09,120.84) .. (59.31,120.84) .. controls (45.54,120.84) and (34.37,111.46) .. (34.37,99.9) -- cycle ;
%Shape: Ellipse [id:dp19170430262813265] 
\draw  [color={rgb, 255:red, 65; green, 117; blue, 5 }  ,draw opacity=1 ][fill={rgb, 255:red, 65; green, 117; blue, 5 }  ,fill opacity=0.18 ] (39.8,103.41) .. controls (39.8,100.54) and (42.97,98.21) .. (46.87,98.21) .. controls (50.78,98.21) and (53.95,100.54) .. (53.95,103.41) .. controls (53.95,106.28) and (50.78,108.6) .. (46.87,108.6) .. controls (42.97,108.6) and (39.8,106.28) .. (39.8,103.41) -- cycle ;
%Shape: Ellipse [id:dp04638687678971076] 
\draw  [color={rgb, 255:red, 65; green, 117; blue, 5 }  ,draw opacity=1 ][fill={rgb, 255:red, 65; green, 117; blue, 5 }  ,fill opacity=0.73 ] (60.22,112.64) .. controls (60.22,109.77) and (63.39,107.44) .. (67.3,107.44) .. controls (71.21,107.44) and (74.37,109.77) .. (74.37,112.64) .. controls (74.37,115.51) and (71.21,117.84) .. (67.3,117.84) .. controls (63.39,117.84) and (60.22,115.51) .. (60.22,112.64) -- cycle ;
%Shape: Ellipse [id:dp6044317617619401] 
\draw  [color={rgb, 255:red, 65; green, 117; blue, 5 }  ,draw opacity=1 ][fill={rgb, 255:red, 65; green, 117; blue, 5 }  ,fill opacity=0.38 ] (53.45,89.17) .. controls (53.45,86.3) and (56.61,83.97) .. (60.52,83.97) .. controls (64.43,83.97) and (67.6,86.3) .. (67.6,89.17) .. controls (67.6,92.04) and (64.43,94.37) .. (60.52,94.37) .. controls (56.61,94.37) and (53.45,92.04) .. (53.45,89.17) -- cycle ;
%Shape: Ellipse [id:dp24081615774375642] 
\draw  [color={rgb, 255:red, 73; green, 10; blue, 86 }  ,draw opacity=1 ][fill={rgb, 255:red, 189; green, 16; blue, 224 }  ,fill opacity=0.33 ] (108.24,68.01) .. controls (108.24,65.14) and (111.4,62.81) .. (115.31,62.81) .. controls (119.22,62.81) and (122.39,65.14) .. (122.39,68.01) .. controls (122.39,70.88) and (119.22,73.21) .. (115.31,73.21) .. controls (111.4,73.21) and (108.24,70.88) .. (108.24,68.01) -- cycle ;
%Shape: Ellipse [id:dp5652262009471116] 
\draw  [color={rgb, 255:red, 67; green, 8; blue, 79 }  ,draw opacity=1 ][fill={rgb, 255:red, 189; green, 16; blue, 224 }  ,fill opacity=0.18 ] (119.78,75.96) .. controls (119.78,73.09) and (122.95,70.76) .. (126.86,70.76) .. controls (130.76,70.76) and (133.93,73.09) .. (133.93,75.96) .. controls (133.93,78.83) and (130.76,81.16) .. (126.86,81.16) .. controls (122.95,81.16) and (119.78,78.83) .. (119.78,75.96) -- cycle ;
%Shape: Ellipse [id:dp6182338948936894] 
\draw  [color={rgb, 255:red, 70; green, 6; blue, 83 }  ,draw opacity=1 ][fill={rgb, 255:red, 189; green, 16; blue, 224 }  ,fill opacity=1 ] (128.07,63.98) .. controls (128.07,61.11) and (131.23,58.79) .. (135.14,58.79) .. controls (139.05,58.79) and (142.21,61.11) .. (142.21,63.98) .. controls (142.21,66.86) and (139.05,69.18) .. (135.14,69.18) .. controls (131.23,69.18) and (128.07,66.86) .. (128.07,63.98) -- cycle ;
%Shape: Ellipse [id:dp7045274719198951] 
\draw  [color={rgb, 255:red, 65; green, 117; blue, 5 }  ,draw opacity=0.6 ][fill={rgb, 255:red, 65; green, 117; blue, 5 }  ,fill opacity=0.27 ] (311.95,102.39) .. controls (324.22,95.58) and (340.74,96.96) .. (348.84,105.46) .. controls (356.94,113.95) and (353.56,126.36) .. (341.28,133.16) .. controls (329.01,139.96) and (312.49,138.58) .. (304.39,130.09) .. controls (296.29,121.59) and (299.67,109.19) .. (311.95,102.39) -- cycle ;
%Shape: Ellipse [id:dp9196558390059527] 
\draw  [color={rgb, 255:red, 65; green, 117; blue, 5 }  ,draw opacity=1 ][fill={rgb, 255:red, 65; green, 117; blue, 5 }  ,fill opacity=0.18 ] (311.42,107.8) .. controls (314.46,106.11) and (318.79,106.7) .. (321.09,109.11) .. controls (323.39,111.52) and (322.78,114.84) .. (319.74,116.53) .. controls (316.69,118.21) and (312.36,117.63) .. (310.06,115.22) .. controls (307.76,112.81) and (308.37,109.49) .. (311.42,107.8) -- cycle ;
%Shape: Ellipse [id:dp9983079787337704] 
\draw  [color={rgb, 255:red, 65; green, 117; blue, 5 }  ,draw opacity=1 ][fill={rgb, 255:red, 65; green, 117; blue, 5 }  ,fill opacity=0.73 ] (313.63,125.83) .. controls (316.68,124.14) and (321.01,124.72) .. (323.3,127.13) .. controls (325.6,129.54) and (324.99,132.86) .. (321.95,134.55) .. controls (318.9,136.24) and (314.57,135.65) .. (312.27,133.24) .. controls (309.98,130.83) and (310.58,127.51) .. (313.63,125.83) -- cycle ;
%Shape: Ellipse [id:dp7529381061622181] 
\draw  [color={rgb, 255:red, 65; green, 117; blue, 5 }  ,draw opacity=1 ][fill={rgb, 255:red, 65; green, 117; blue, 5 }  ,fill opacity=0.38 ] (334.55,107.85) .. controls (337.59,106.16) and (341.92,106.74) .. (344.22,109.15) .. controls (346.52,111.56) and (345.91,114.89) .. (342.87,116.57) .. controls (339.82,118.26) and (335.49,117.68) .. (333.19,115.27) .. controls (330.89,112.86) and (331.5,109.53) .. (334.55,107.85) -- cycle ;
%Shape: Ellipse [id:dp46177239016470617] 
\draw  [color={rgb, 255:red, 73; green, 10; blue, 86 }  ,draw opacity=1 ][fill={rgb, 255:red, 189; green, 16; blue, 224 }  ,fill opacity=0.33 ] (404.92,55.49) .. controls (407.97,53.8) and (412.3,54.38) .. (414.59,56.79) .. controls (416.89,59.2) and (416.28,62.53) .. (413.24,64.21) .. controls (410.19,65.9) and (405.86,65.32) .. (403.56,62.91) .. controls (401.27,60.5) and (401.87,57.17) .. (404.92,55.49) -- cycle ;
%Shape: Ellipse [id:dp3131255331682834] 
\draw  [color={rgb, 255:red, 67; green, 8; blue, 79 }  ,draw opacity=1 ][fill={rgb, 255:red, 189; green, 16; blue, 224 }  ,fill opacity=0.18 ] (403.27,67.28) .. controls (406.32,65.59) and (410.65,66.18) .. (412.94,68.59) .. controls (415.24,71) and (414.64,74.32) .. (411.59,76.01) .. controls (408.54,77.7) and (404.21,77.11) .. (401.91,74.7) .. controls (399.62,72.29) and (400.22,68.97) .. (403.27,67.28) -- cycle ;
%Shape: Ellipse [id:dp4151992664711134] 
\draw  [color={rgb, 255:red, 70; green, 6; blue, 83 }  ,draw opacity=1 ][fill={rgb, 255:red, 189; green, 16; blue, 224 }  ,fill opacity=1 ] (420.85,65.35) .. controls (423.89,63.66) and (428.23,64.25) .. (430.52,66.66) .. controls (432.82,69.07) and (432.21,72.39) .. (429.17,74.08) .. controls (426.12,75.76) and (421.79,75.18) .. (419.49,72.77) .. controls (417.2,70.36) and (417.8,67.04) .. (420.85,65.35) -- cycle ;
%Shape: Ellipse [id:dp030329347446797494] 
\draw  [color={rgb, 255:red, 187; green, 131; blue, 38 }  ,draw opacity=0.28 ][fill={rgb, 255:red, 245; green, 166; blue, 35 }  ,fill opacity=0.25 ] (39.2,104.2) .. controls (39.2,92.63) and (50.37,83.25) .. (64.15,83.25) .. controls (77.93,83.25) and (89.09,92.63) .. (89.09,104.2) .. controls (89.09,115.76) and (77.93,125.14) .. (64.15,125.14) .. controls (50.37,125.14) and (39.2,115.76) .. (39.2,104.2) -- cycle ;
%Shape: Ellipse [id:dp712129807013384] 
\draw  [color={rgb, 255:red, 187; green, 131; blue, 38 }  ,draw opacity=0.93 ][fill={rgb, 255:red, 245; green, 166; blue, 35 }  ,fill opacity=1 ] (44.63,107.71) .. controls (44.63,104.84) and (47.8,102.51) .. (51.71,102.51) .. controls (55.61,102.51) and (58.78,104.84) .. (58.78,107.71) .. controls (58.78,110.58) and (55.61,112.9) .. (51.71,112.9) .. controls (47.8,112.9) and (44.63,110.58) .. (44.63,107.71) -- cycle ;
%Shape: Ellipse [id:dp680235146004749] 
\draw  [color={rgb, 255:red, 187; green, 131; blue, 38 }  ,draw opacity=0.93 ][fill={rgb, 255:red, 245; green, 166; blue, 35 }  ,fill opacity=0.68 ] (67.48,114.18) .. controls (67.48,111.3) and (70.64,108.98) .. (74.55,108.98) .. controls (78.46,108.98) and (81.63,111.3) .. (81.63,114.18) .. controls (81.63,117.05) and (78.46,119.37) .. (74.55,119.37) .. controls (70.64,119.37) and (67.48,117.05) .. (67.48,114.18) -- cycle ;
%Shape: Ellipse [id:dp048333949216573746] 
\draw  [color={rgb, 255:red, 187; green, 131; blue, 38 }  ,draw opacity=0.93 ][fill={rgb, 255:red, 245; green, 166; blue, 35 }  ,fill opacity=0.45 ] (57.07,99) .. controls (57.07,96.13) and (60.24,93.8) .. (64.15,93.8) .. controls (68.06,93.8) and (71.22,96.13) .. (71.22,99) .. controls (71.22,101.87) and (68.06,104.2) .. (64.15,104.2) .. controls (60.24,104.2) and (57.07,101.87) .. (57.07,99) -- cycle ;
%Shape: Ellipse [id:dp3523034877521757] 
\draw  [color={rgb, 255:red, 57; green, 75; blue, 100 }  ,draw opacity=1 ][fill={rgb, 255:red, 74; green, 144; blue, 226 }  ,fill opacity=0.42 ] (113.07,72.31) .. controls (113.07,69.44) and (116.24,67.11) .. (120.15,67.11) .. controls (124.05,67.11) and (127.22,69.44) .. (127.22,72.31) .. controls (127.22,75.18) and (124.05,77.51) .. (120.15,77.51) .. controls (116.24,77.51) and (113.07,75.18) .. (113.07,72.31) -- cycle ;
%Shape: Ellipse [id:dp7600785742309424] 
\draw  [color={rgb, 255:red, 57; green, 75; blue, 100 }  ,draw opacity=1 ][fill={rgb, 255:red, 74; green, 144; blue, 226 }  ,fill opacity=1 ] (116.51,90.26) .. controls (116.51,87.39) and (119.68,85.06) .. (123.58,85.06) .. controls (127.49,85.06) and (130.66,87.39) .. (130.66,90.26) .. controls (130.66,93.13) and (127.49,95.46) .. (123.58,95.46) .. controls (119.68,95.46) and (116.51,93.13) .. (116.51,90.26) -- cycle ;
%Shape: Ellipse [id:dp5524364703634363] 
\draw  [color={rgb, 255:red, 57; green, 75; blue, 100 }  ,draw opacity=1 ][fill={rgb, 255:red, 74; green, 144; blue, 226 }  ,fill opacity=0.65 ] (131.69,80.26) .. controls (131.69,77.39) and (134.86,75.06) .. (138.77,75.06) .. controls (142.67,75.06) and (145.84,77.39) .. (145.84,80.26) .. controls (145.84,83.13) and (142.67,85.46) .. (138.77,85.46) .. controls (134.86,85.46) and (131.69,83.13) .. (131.69,80.26) -- cycle ;
%Shape: Ellipse [id:dp49442779744950793] 
\draw  [color={rgb, 255:red, 74; green, 144; blue, 226 }  ,draw opacity=0.42 ][fill={rgb, 255:red, 74; green, 144; blue, 226 }  ,fill opacity=0.19 ] (106.74,80.26) .. controls (106.74,68.69) and (117.91,59.32) .. (131.69,59.32) .. controls (145.47,59.32) and (156.64,68.69) .. (156.64,80.26) .. controls (156.64,91.83) and (145.47,101.21) .. (131.69,101.21) .. controls (117.91,101.21) and (106.74,91.83) .. (106.74,80.26) -- cycle ;
%Straight Lines [id:da9604265155419616] 
\draw [color={rgb, 255:red, 74; green, 74; blue, 74 }  ,draw opacity=1 ] [dash pattern={on 4.5pt off 4.5pt}]  (48,43.42) -- (157.33,146.04) ;
%Shape: Ellipse [id:dp5621672568885412] 
\draw  [color={rgb, 255:red, 155; green, 155; blue, 155 }  ,draw opacity=1 ][fill={rgb, 255:red, 155; green, 155; blue, 155 }  ,fill opacity=0.28 ] (241.03,138.24) .. controls (241.03,126.67) and (252.2,117.29) .. (265.97,117.29) .. controls (279.75,117.29) and (290.92,126.67) .. (290.92,138.24) .. controls (290.92,149.81) and (279.75,159.19) .. (265.97,159.19) .. controls (252.2,159.19) and (241.03,149.81) .. (241.03,138.24) -- cycle ;
%Shape: Ellipse [id:dp0472948541463275] 
\draw  [color={rgb, 255:red, 155; green, 155; blue, 155 }  ,draw opacity=1 ][fill={rgb, 255:red, 155; green, 155; blue, 155 }  ,fill opacity=0.58 ] (246.46,128.85) .. controls (246.46,125.98) and (249.62,123.65) .. (253.53,123.65) .. controls (257.44,123.65) and (260.61,125.98) .. (260.61,128.85) .. controls (260.61,131.72) and (257.44,134.05) .. (253.53,134.05) .. controls (249.62,134.05) and (246.46,131.72) .. (246.46,128.85) -- cycle ;
%Shape: Ellipse [id:dp7271329189270219] 
\draw  [color={rgb, 255:red, 128; green, 128; blue, 128 }  ,draw opacity=1 ][fill={rgb, 255:red, 155; green, 155; blue, 155 }  ,fill opacity=0.22 ] (246.34,144.53) .. controls (246.34,141.66) and (249.51,139.34) .. (253.41,139.34) .. controls (257.32,139.34) and (260.49,141.66) .. (260.49,144.53) .. controls (260.49,147.4) and (257.32,149.73) .. (253.41,149.73) .. controls (249.51,149.73) and (246.34,147.4) .. (246.34,144.53) -- cycle ;
%Shape: Ellipse [id:dp6762807226541048] 
\draw  [color={rgb, 255:red, 155; green, 155; blue, 155 }  ,draw opacity=1 ][fill={rgb, 255:red, 155; green, 155; blue, 155 }  ,fill opacity=0.77 ] (276.77,138.24) .. controls (276.77,135.37) and (279.94,133.04) .. (283.84,133.04) .. controls (287.75,133.04) and (290.92,135.37) .. (290.92,138.24) .. controls (290.92,141.11) and (287.75,143.44) .. (283.84,143.44) .. controls (279.94,143.44) and (276.77,141.11) .. (276.77,138.24) -- cycle ;
%Shape: Ellipse [id:dp9235237162293723] 
\draw  [color={rgb, 255:red, 132; green, 8; blue, 23 }  ,draw opacity=1 ][fill={rgb, 255:red, 208; green, 2; blue, 27 }  ,fill opacity=0.67 ] (309.83,28.46) .. controls (309.83,25.42) and (312.9,22.96) .. (316.69,22.96) .. controls (320.48,22.96) and (323.55,25.42) .. (323.55,28.46) .. controls (323.55,31.49) and (320.48,33.96) .. (316.69,33.96) .. controls (312.9,33.96) and (309.83,31.49) .. (309.83,28.46) -- cycle ;
%Shape: Ellipse [id:dp3891273259453849] 
\draw  [color={rgb, 255:red, 129; green, 8; blue, 23 }  ,draw opacity=1 ][fill={rgb, 255:red, 208; green, 2; blue, 27 }  ,fill opacity=1 ] (292.58,46.91) .. controls (292.58,44.04) and (295.75,41.71) .. (299.66,41.71) .. controls (303.56,41.71) and (306.73,44.04) .. (306.73,46.91) .. controls (306.73,49.78) and (303.56,52.11) .. (299.66,52.11) .. controls (295.75,52.11) and (292.58,49.78) .. (292.58,46.91) -- cycle ;
%Shape: Ellipse [id:dp2683840207885593] 
\draw  [color={rgb, 255:red, 125; green, 9; blue, 24 }  ,draw opacity=1 ][fill={rgb, 255:red, 208; green, 2; blue, 27 }  ,fill opacity=0.13 ] (312.6,44.28) .. controls (312.6,41.41) and (315.76,39.08) .. (319.67,39.08) .. controls (323.58,39.08) and (326.75,41.41) .. (326.75,44.28) .. controls (326.75,47.15) and (323.58,49.48) .. (319.67,49.48) .. controls (315.76,49.48) and (312.6,47.15) .. (312.6,44.28) -- cycle ;
%Shape: Ellipse [id:dp2241475837877669] 
\draw  [color={rgb, 255:red, 208; green, 2; blue, 27 }  ,draw opacity=0.74 ][fill={rgb, 255:red, 208; green, 2; blue, 27 }  ,fill opacity=0.22 ] (111.58,85.82) .. controls (111.58,74.25) and (122.75,64.87) .. (136.52,64.87) .. controls (150.3,64.87) and (161.47,74.25) .. (161.47,85.82) .. controls (161.47,97.39) and (150.3,106.77) .. (136.52,106.77) .. controls (122.75,106.77) and (111.58,97.39) .. (111.58,85.82) -- cycle ;
%Shape: Ellipse [id:dp12280146645788359] 
\draw  [color={rgb, 255:red, 155; green, 155; blue, 155 }  ,draw opacity=1 ][fill={rgb, 255:red, 155; green, 155; blue, 155 }  ,fill opacity=0.28 ] (44.04,109.75) .. controls (44.04,98.19) and (55.2,88.81) .. (68.98,88.81) .. controls (82.76,88.81) and (93.93,98.19) .. (93.93,109.75) .. controls (93.93,121.32) and (82.76,130.7) .. (68.98,130.7) .. controls (55.2,130.7) and (44.04,121.32) .. (44.04,109.75) -- cycle ;
%Shape: Ellipse [id:dp1779389444680003] 
\draw  [color={rgb, 255:red, 155; green, 155; blue, 155 }  ,draw opacity=1 ][fill={rgb, 255:red, 155; green, 155; blue, 155 }  ,fill opacity=0.58 ] (49.47,100.37) .. controls (49.47,97.5) and (52.63,95.17) .. (56.54,95.17) .. controls (60.45,95.17) and (63.62,97.5) .. (63.62,100.37) .. controls (63.62,103.24) and (60.45,105.56) .. (56.54,105.56) .. controls (52.63,105.56) and (49.47,103.24) .. (49.47,100.37) -- cycle ;
%Shape: Ellipse [id:dp41282473036148726] 
\draw  [color={rgb, 255:red, 128; green, 128; blue, 128 }  ,draw opacity=1 ][fill={rgb, 255:red, 155; green, 155; blue, 155 }  ,fill opacity=0.22 ] (49.35,116.05) .. controls (49.35,113.18) and (52.52,110.85) .. (56.42,110.85) .. controls (60.33,110.85) and (63.5,113.18) .. (63.5,116.05) .. controls (63.5,118.92) and (60.33,121.25) .. (56.42,121.25) .. controls (52.52,121.25) and (49.35,118.92) .. (49.35,116.05) -- cycle ;
%Shape: Ellipse [id:dp19873833089399318] 
\draw  [color={rgb, 255:red, 155; green, 155; blue, 155 }  ,draw opacity=1 ][fill={rgb, 255:red, 155; green, 155; blue, 155 }  ,fill opacity=0.77 ] (79.78,109.75) .. controls (79.78,106.88) and (82.95,104.56) .. (86.85,104.56) .. controls (90.76,104.56) and (93.93,106.88) .. (93.93,109.75) .. controls (93.93,112.63) and (90.76,114.95) .. (86.85,114.95) .. controls (82.95,114.95) and (79.78,112.63) .. (79.78,109.75) -- cycle ;
%Shape: Ellipse [id:dp4383938551326363] 
\draw  [color={rgb, 255:red, 132; green, 8; blue, 23 }  ,draw opacity=1 ][fill={rgb, 255:red, 208; green, 2; blue, 27 }  ,fill opacity=0.67 ] (129.45,70.07) .. controls (129.45,67.2) and (132.62,64.87) .. (136.52,64.87) .. controls (140.43,64.87) and (143.6,67.2) .. (143.6,70.07) .. controls (143.6,72.94) and (140.43,75.27) .. (136.52,75.27) .. controls (132.62,75.27) and (129.45,72.94) .. (129.45,70.07) -- cycle ;
%Shape: Ellipse [id:dp16066701265363958] 
\draw  [color={rgb, 255:red, 129; green, 8; blue, 23 }  ,draw opacity=1 ][fill={rgb, 255:red, 208; green, 2; blue, 27 }  ,fill opacity=1 ] (121.34,95.82) .. controls (121.34,92.95) and (124.51,90.62) .. (128.42,90.62) .. controls (132.32,90.62) and (135.49,92.95) .. (135.49,95.82) .. controls (135.49,98.69) and (132.32,101.02) .. (128.42,101.02) .. controls (124.51,101.02) and (121.34,98.69) .. (121.34,95.82) -- cycle ;
%Shape: Ellipse [id:dp9196637935627724] 
\draw  [color={rgb, 255:red, 125; green, 9; blue, 24 }  ,draw opacity=1 ][fill={rgb, 255:red, 208; green, 2; blue, 27 }  ,fill opacity=0.13 ] (141.36,93.19) .. controls (141.36,90.32) and (144.53,87.99) .. (148.43,87.99) .. controls (152.34,87.99) and (155.51,90.32) .. (155.51,93.19) .. controls (155.51,96.06) and (152.34,98.39) .. (148.43,98.39) .. controls (144.53,98.39) and (141.36,96.06) .. (141.36,93.19) -- cycle ;
%Shape: Ellipse [id:dp8123672363209512] 
\draw  [color={rgb, 255:red, 73; green, 10; blue, 86 }  ,draw opacity=1 ][fill={rgb, 255:red, 189; green, 16; blue, 224 }  ,fill opacity=0.33 ] (560.92,82.49) .. controls (563.97,80.8) and (568.3,81.38) .. (570.59,83.79) .. controls (572.89,86.2) and (572.28,89.53) .. (569.24,91.21) .. controls (566.19,92.9) and (561.86,92.32) .. (559.56,89.91) .. controls (557.27,87.5) and (557.87,84.17) .. (560.92,82.49) -- cycle ;
%Shape: Ellipse [id:dp2570511302396914] 
\draw  [color={rgb, 255:red, 65; green, 117; blue, 5 }  ,draw opacity=1 ][fill={rgb, 255:red, 65; green, 117; blue, 5 }  ,fill opacity=0.73 ] (585.63,83.83) .. controls (588.68,82.14) and (593.01,82.72) .. (595.3,85.13) .. controls (597.6,87.54) and (596.99,90.86) .. (593.95,92.55) .. controls (590.9,94.24) and (586.57,93.65) .. (584.27,91.24) .. controls (581.98,88.83) and (582.58,85.51) .. (585.63,83.83) -- cycle ;
%Shape: Ellipse [id:dp6181750375276943] 
\draw  [color={rgb, 255:red, 132; green, 8; blue, 23 }  ,draw opacity=1 ][fill={rgb, 255:red, 208; green, 2; blue, 27 }  ,fill opacity=0.67 ] (606.83,108.46) .. controls (606.83,105.42) and (609.9,102.96) .. (613.69,102.96) .. controls (617.48,102.96) and (620.55,105.42) .. (620.55,108.46) .. controls (620.55,111.49) and (617.48,113.96) .. (613.69,113.96) .. controls (609.9,113.96) and (606.83,111.49) .. (606.83,108.46) -- cycle ;
%Shape: Ellipse [id:dp43417455910667946] 
\draw  [color={rgb, 255:red, 155; green, 155; blue, 155 }  ,draw opacity=1 ][fill={rgb, 255:red, 155; green, 155; blue, 155 }  ,fill opacity=0.58 ] (629.46,108.85) .. controls (629.46,105.98) and (632.62,103.65) .. (636.53,103.65) .. controls (640.44,103.65) and (643.61,105.98) .. (643.61,108.85) .. controls (643.61,111.72) and (640.44,114.05) .. (636.53,114.05) .. controls (632.62,114.05) and (629.46,111.72) .. (629.46,108.85) -- cycle ;
%Shape: Ellipse [id:dp7320177694846334] 
\draw  [color={rgb, 255:red, 187; green, 131; blue, 38 }  ,draw opacity=0.93 ][fill={rgb, 255:red, 245; green, 166; blue, 35 }  ,fill opacity=0.45 ] (507.44,66.18) .. controls (507.44,63.31) and (510.61,60.98) .. (514.52,60.98) .. controls (518.43,60.98) and (521.59,63.31) .. (521.59,66.18) .. controls (521.59,69.05) and (518.43,71.37) .. (514.52,71.37) .. controls (510.61,71.37) and (507.44,69.05) .. (507.44,66.18) -- cycle ;
%Shape: Ellipse [id:dp48048051407271986] 
\draw  [color={rgb, 255:red, 57; green, 75; blue, 100 }  ,draw opacity=1 ][fill={rgb, 255:red, 74; green, 144; blue, 226 }  ,fill opacity=0.65 ] (533.06,65.44) .. controls (533.06,62.57) and (536.23,60.24) .. (540.14,60.24) .. controls (544.04,60.24) and (547.21,62.57) .. (547.21,65.44) .. controls (547.21,68.31) and (544.04,70.64) .. (540.14,70.64) .. controls (536.23,70.64) and (533.06,68.31) .. (533.06,65.44) -- cycle ;

% Text Node
\draw (255,57.66) node [anchor=north west][inner sep=0.75pt]  [font=\scriptsize,color={rgb, 255:red, 245; green, 166; blue, 35 }  ,opacity=1 ] [align=left] {Class A};
% Text Node
\draw (357.94,88.66) node [anchor=north west][inner sep=0.75pt]  [font=\scriptsize,color={rgb, 255:red, 74; green, 144; blue, 226 }  ,opacity=1 ,rotate=-24.08] [align=left] {Class B};
% Text Node
\draw (52.46,183.92) node [anchor=north west][inner sep=0.75pt]  [font=\normalsize] [align=left] {{\fontfamily{ptm}\selectfont \textbf{\textcolor[rgb]{0.29,0.29,0.29}{Existing Methods}}}};
% Text Node
\draw (271.98,182.92) node [anchor=north west][inner sep=0.75pt]  [font=\normalsize] [align=left] {{\fontfamily{ptm}\selectfont \textcolor[rgb]{0.29,0.29,0.29}{\textbf{Proposed \ Method}}}};
% Text Node
\draw (24.27,69.1) node [anchor=north west][inner sep=0.75pt]  [font=\scriptsize,color={rgb, 255:red, 65; green, 117; blue, 5 }  ,opacity=1 ] [align=left] {Class D};
% Text Node
\draw (163.12,82.59) node [anchor=north west][inner sep=0.75pt]  [font=\scriptsize,color={rgb, 255:red, 189; green, 16; blue, 224 }  ,opacity=1 ] [align=left] {Class C};
% Text Node
\draw (296.91,118.38) node [anchor=north west][inner sep=0.75pt]  [font=\scriptsize,color={rgb, 255:red, 65; green, 117; blue, 5 }  ,opacity=1 ,rotate=-53.99] [align=left] {Class D};
% Text Node
\draw (409.54,89.11) node [anchor=north west][inner sep=0.75pt]  [font=\scriptsize,color={rgb, 255:red, 189; green, 16; blue, 224 }  ,opacity=1 ,rotate=-53.99] [align=left] {Class C};
% Text Node
\draw (27.79,60.5) node [anchor=north west][inner sep=0.75pt]  [font=\scriptsize,color={rgb, 255:red, 245; green, 166; blue, 35 }  ,opacity=1 ] [align=left] {Class A};
% Text Node
\draw (147.74,95.52) node [anchor=north west][inner sep=0.75pt]  [font=\scriptsize,color={rgb, 255:red, 74; green, 144; blue, 226 }  ,opacity=1 ] [align=left] {Class B};
% Text Node
\draw (285.21,151.67) node [anchor=north west][inner sep=0.75pt]  [font=\scriptsize,color={rgb, 255:red, 155; green, 155; blue, 155 }  ,opacity=1 ] [align=left] {Class F};
% Text Node
\draw (331.22,24.6) node [anchor=north west][inner sep=0.75pt]  [font=\scriptsize,color={rgb, 255:red, 208; green, 2; blue, 27 }  ,opacity=1 ] [align=left] {Class E};
% Text Node
\draw (53.17,130.55) node [anchor=north west][inner sep=0.75pt]  [font=\scriptsize,color={rgb, 255:red, 155; green, 155; blue, 155 }  ,opacity=1 ] [align=left] {Class F};
% Text Node
\draw (142.9,109.1) node [anchor=north west][inner sep=0.75pt]  [font=\scriptsize,color={rgb, 255:red, 208; green, 2; blue, 27 }  ,opacity=1 ] [align=left] {Class E};
% Text Node
\draw (454,15) node [anchor=north west][inner sep=0.75pt]   [align=left] {{\fontfamily{ptm}\selectfont \hspace{2.5cm} Data}\\{\fontfamily{ptm}\selectfont Label \ \ \ A \ \ B \ \ C \ \ D \ \ E \ \ F}\\{\fontfamily{ptm}\selectfont Client1 }\\{\fontfamily{ptm}\selectfont Client2 }\\{\fontfamily{ptm}\selectfont Client3}};

\end{tikzpicture}
    \caption{The Representational Collapse Problem in Federated Learning. Under extreme data heterogeneity (e.g., Client 1: $\{a,b\}$, Client 2: $\{c,d\}$, Client 3: $\{e,f\}$), local models optimise feature spaces without knowledge of global class boundaries. While local embeddings remain well-separated, the absence of shared negative samples results in conflicting coordinate systems. Upon model aggregation, these discrepancies cause the global embedding space to collapse, destroying inter-class division and discriminative structure.}
    \label{fig:problem}
\end{figure*}
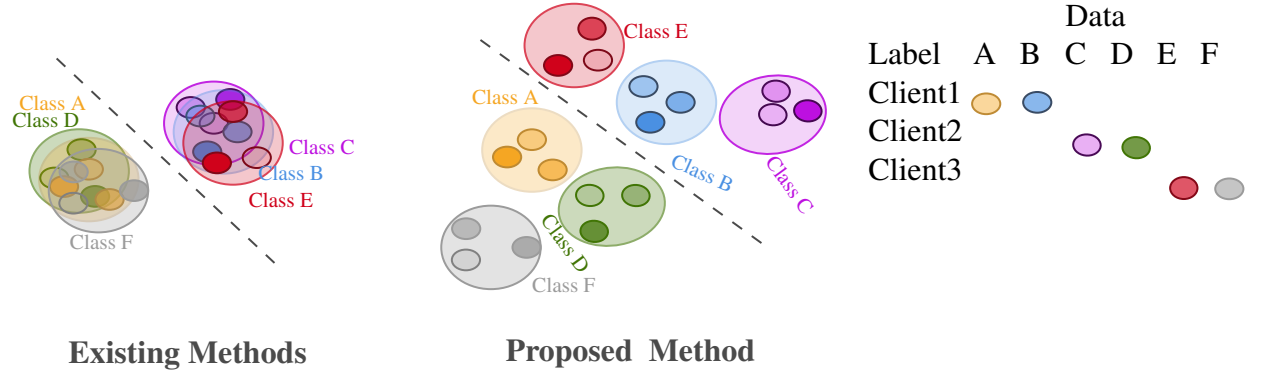

Data decentralisation presents a significant challenge for the deployment and scalability of deep learning models, as raw data cannot always be centralised due to accessibility, privacy, or regulatory constraints. Privacy concerns are particularly sensitive, as many organisations are hesitant to share local datasets, or storing in one server, for confidentiality reasons. To address these issues, Federated Learning (FL) has emerged as a promising paradigm, enabling collaborative model training without exposing sensitive raw data \cite{mora2024enhancing, zhu2021federated}. The pioneering FedAvg algorithm \cite{fedavg} operates by maintaining local datasets on clients while transmitting only model updates to a central server. This approach preserves data in clients server while allowing collaborative learning across distributed nodes. FL has proven essential for high-impact applications, including medical imaging, remote sensing, and large-scale image classification, where both accessibility and decentralisation are critical, \cite{teo2024federated}, \cite{lin2025fedrsclip}.

Statistical heterogeneity of local client data remains a key challenge in federated learning. In practice, client datasets are rarely identically distributed (non-IID), exhibiting variations in class proportions, sample sizes, and feature distributions. Such heterogeneity complicates the training of a global model that generalises effectively across all clients. Therefore, the performance of FL algorithms can degrade substantially under limited local data or severe class imbalance, both of which are common in real-world scenarios.

A major consequence of this heterogeneity is \textbf{representational collapse}, where the feature space loses discriminative capacity, and embeddings from different classes or clients become indistinguishable. As illustrated in Figure \ref{fig:problem}, divergent local training leads to inconsistent feature representations across clients. Aggregation of these misaligned representations at the central server can cause the global model to collapse, resulting in overlapping or trivial embeddings that impair generalisation and weaken representational robustness.

Several approaches have been proposed to mitigate data heterogeneity during local training in federated learning \cite{vahidian2023rethinking}, \cite{wangnew}, \cite{xma}. FedProx \cite{fedprox} augments the local objective with a proximal term that penalises deviations between local and global model parameters, ensuring smoother and more stable updates. SCAFFOLD \cite{karimireddy2019scaffold} addresses client drift by introducing control variates to correct gradient dissimilarities across clients. These methods highlight that incorporating global model knowledge can improve the robustness of local representations.

Beyond regularisation-based techniques, contrastive learning has emerged as a powerful strategy for leveraging global knowledge in FL \cite{Fedproc}. MOON \cite{li2021model} mitigates data heterogeneity by maximising agreement between local and global model embeddings, showing how global knowledge enhances local robustness. Similarly, FedRCL \cite{seo2024relaxed} enforces inter-class variance in the feature space, ensuring that data points of the same class remain well-clustered. This separation preserves representational diversity, enabling more effective learning and improved generalisation.

While aligning local and global representations is critical for effective federated learning, overly aggressive alignment can lead to representational collapse, where intra-class diversity within client models is reduced. This collapse undermines the model’s ability to capture subtle variations within the same class, ultimately degrading both local and global performance. Addressing this issue requires a careful balance between intra-class minimal variance and maximum inter-class variance, ensuring that learned representations remain both discriminative and diverse across clients. This motivates the following key question:

\textit{“How can we preserve both intra-class and inter-class relationships to prevent representational collapse in global and local models under heterogeneous data?”}

We propose FedQuad \footnote{This article is an extended version of our conference paper published at FLTA 2025 \cite{goksu2025fedquad}, with additional experiments, non-IID analyses, and ablation studies.}, a metric learning-based federated learning approach designed to mitigate representational collapse under heterogeneous data. Unlike conventional contrastive or triplet-based methods \cite{khosla2020supervised, hoffer2015deep}, FedQuad explicitly models relative distances between samples using stochastic quadruplets, consisting of an anchor, a positive (same class), a negative (different class), and a harder negative (also from a different class). The proposed loss simultaneously minimises the distance between the anchor and positive while maximising distances to both negatives. This encourages a feature space where intra-class samples are tightly clustered, and inter-class samples remain well-separated. By preserving both intra-class diversity and inter-class discriminability, FedQuad generates robust representations that are resilient to aggregation-induced degradation, directly addressing the core challenges of representational collapse in federated learning.\\
To summarise, our major contributions are as follows:

\begin{itemize}
    \item We propose a novel metric learning-based federated learning framework to address representational collapse under heterogeneous client data.
    \item We analyse the impact of intra-class and inter-class variance, highlighting their roles in preserving discriminative representations across clients.
    \item We introduce a quadruplet-based loss function that effectively mitigates representational collapse in both local and global models.
    \item We design an offline stochastic quadruplet sampling strategy per client, tailored for imbalanced and non-IID data distributions, ensuring robust and diverse training.
    \item We demonstrate the effectiveness of our method under scenarios with a large number of clients, using random client selection and minimal participation per round.
\end{itemize}
Traditional quadruplet loss \cite{chen2017beyond} primarily minimises the distance between the anchor and positive sample, providing only a weak push against negative samples. Typically, it enforces a margin for a single negative pair, offering limited guidance for handling multiple or harder negatives. Consequently, its effectiveness is reduced in highly heterogeneous settings such as federated learning, where negative samples can vary widely in difficulty. Without explicitly modelling strong separation from challenging negatives, learned representations may lack sufficient discriminative structure, compromising their ability to preserve inter-class boundaries.

Hard negative mining plays a critical role in contrastive and triplet-based methods \cite{xuan2020hard}. In FedQuad, we introduce a class-separation-aware quadruplet construction strategy that ensures each batch contains at least one positive pair and negative samples drawn from different classes than the anchor. This design provides more informative and relevant optimisation, particularly in federated contexts with heterogeneous client data distributions. Moreover, our method improves representation learning for underrepresented classes, where conventional contrastive losses often fail due to insufficient informative negatives.

\section{Background}

\subsection{Federated Learning}
FL frameworks such as FedAvg typically involve three main steps: \textit{broadcasting}, \textit{local training}, and \textit{model aggregation}. After each communication round, the central server broadcasts the current global model to all participating clients. Each client then performs local training on its private data, ensuring data privacy by keeping raw data on-device. Once local updates are completed (e.g., after a fixed number of epochs), clients send their updated model parameters back to the server. The server performs model aggregation, typically by averaging the received parameters, to form a new global model, which is then broadcast in the next round.
Following the typical FL approach, the data $D$ is distributed across $K$ clients. Let $D_k$ be the local dataset at client $k$, with $n_k = |D_k|$ denoting the number of data points at client $k$. The global objective function in FL is then a weighted average of the local objectives
\begin{equation}
\min _w F(w)=\sum_{k=1}^K \frac{n_k}{n} F_k(w)
\end{equation}
where $n = \sum_{k=1}^{K} n_k$ is the total number of data points across all clients and $F_k(w)$ is the local objective function at a client $k$
\begin{equation}
 \quad F_k(w)=\frac{1}{n_k} \sum_{i \in \mathcal{D}_k} f_i(w)
\end{equation}
To the best of our knowledge, this is the first work to systematically explore and analyse metric learning losses for representation learning in federated learning, with a focus on local client updates under heterogeneous data. Prior research has primarily addressed data heterogeneity through two complementary strategies: improving local training or optimising global aggregation. Our approach falls into the local training improvement category. \\

\textbf{Federated Metric Learning} \\
Existing federated metric learning methods, such as FedMetric \cite{park2021federated}, learn embeddings using a metric loss that treats positive and negative pairs differently. However, these approaches often fail to explicitly model the relative similarity between a positive sample and multiple negative samples, limiting their ability to capture fine-grained structures in the representation space. Moreover, FedMetric relies on proxy-based hypersphere clustering, which oversimplifies the underlying data distribution. Other works \cite{tian2022privacy, gu2023defending, shao2023privacy} focus on designing or enhancing federated metric learning models, rather than explicitly applying metric learning losses (e.g., contrastive, triplet, or quadruplet loss) during local client training.

While metric learning has demonstrated effectiveness in supervised representation learning by minimising intra-class variance and maximising inter-class variance ; including Quadruplet loss \cite{chen2017beyond}, Triplet loss \cite{hoffer2015deep}, and supervised contrastive loss \cite{khosla2020supervised}, its integration into federated learning remains largely unexplored. Our work is among the first to systematically investigate metric learning losses in the federated setting, with a focus on preventing representational collapse under severe data heterogeneity. \\

\textbf{Data heterogeneity in Federated Learning} \\ 
A central challenge in federated learning (FL) is the inherently non-IID nature of data across clients. In practice, clients often exhibit significant class imbalance or domain-specific biases in their local datasets, which can substantially hinder both the convergence and generalization performance of the global model. To address these challenges, several techniques have been proposed \cite{gao2022feddc, feddyn, fang2025robust}.

Contrastive learning-based approaches have gained considerable attention in FL for their effectiveness in aligning global and local representations, mitigating client drift and data heterogeneity. For example, FedProc \cite{Fedproc}, FedCRL \cite{huang2024fedcrl}, and FedPCL \cite{tan2022federated} reformulate contrastive objectives to reduce divergence between local and global models, thereby improving alignment. Relaxed contrastive methods such as MOON \cite{li2021model}, FedRCL \cite{seo2024relaxed}, and FedTrip \cite{li2023fedtrip} employ model-level alignment and distribution-aware aggregation to alleviate the impact of heterogeneous data. Specifically, MOON introduces a model-contrastive loss that aligns the current local model with the global model while pushing it away from the previous local model.

Unsupervised contrastive learning techniques \cite{fedu}, \cite{orchestra} including FedSimCLR \cite{louizos2024mutual} and FedMoCo \cite{dong2021federated}, have also been explored in federated settings, leveraging mutual information maximization without requiring labeled data. However, these approaches fall outside the scope of our work, which focuses on supervised image classification with labeled client datasets. In contrast to prior methods, our framework does not rely on an additional global model for contrastive learning, nor does it depend on global prototypes to correct deviations in local training.\\

\section{Methodogy}
\label{sec:3}
\begin{figure*}[htp]
    \centering
    \includegraphics[width=0.9\linewidth]{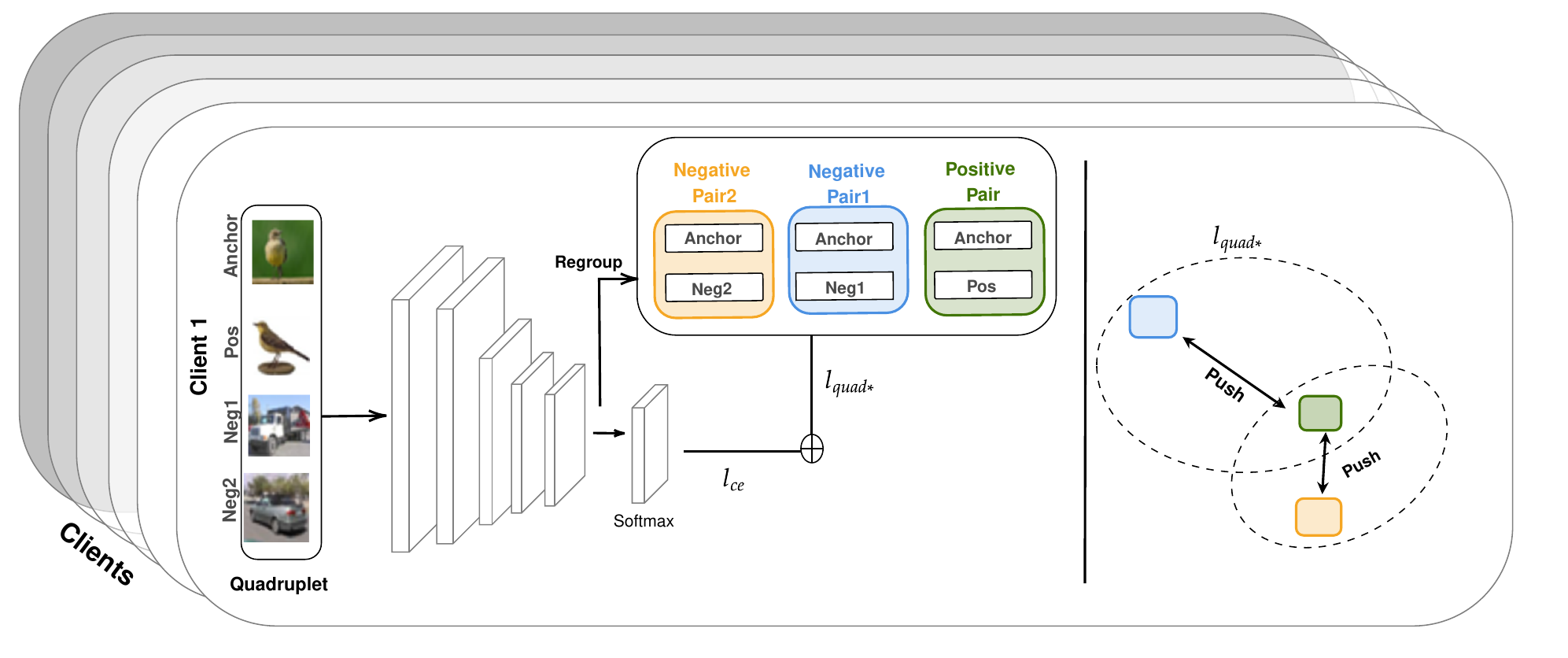}
    \caption{Overview of the FedQuad local training framework. Each client minimises loss composed of the cross-entropy loss $\ell_{ce}$, computed after the softmax layer, and the proposed quadruplet loss $\ell_{quad*}$, applied to the non-normalised embeddings from the encoder. The images are from the CIFAR-10 dataset, which explains the low resolution.}
    \label{fig:method}
\end{figure*}

Our proposed method builds upon the quadruplet loss framework for local training. Unlike the standard quadruplet loss, which includes a negative pair distance term ($n_1, n_2$) to constrain the distance between two negative samples, we focus on modelling the anchor’s relationship with multiple negatives drawn from different classes. In traditional quadruplet loss, the negative pair term often provides only a weak push and contributes minimally to discriminative representation learning. Instead of explicitly modelling the structure among negatives, our approach encourages the positive sample to be well-separated from the entire set of negatives collectively.

Since negative samples originate from different classes, promoting broad separation from the anchor, rather than focusing solely on hard negatives, enhances the discriminative capacity of the learned representations. Our loss function is designed to optimise two key objectives: (i) minimising intra-class variance, ensuring features from the same class remain clustered in the embedding space; and (ii) maximising inter-class separation by simultaneously pushing the anchor away from multiple negatives. This formulation facilitates effective feature separation both at the local client level and in the globally aggregated model.

\subsection{Stochastic Quadruplet Sampling}
We design a stochastic pair sampling strategy to generate quadruplet samples for representation learning. Each quadruplet consists of an anchor, a positive sample drawn from the same class as the anchor, and two negative samples selected from different classes, ensuring that the negatives differ both from the anchor’s class and from each other. To enable efficient sampling, a mapping is first constructed between class labels and sample indices within the provided subset.

During batch generation, a positive sample is randomly chosen from the anchor’s class, ensuring it is distinct from the anchor itself. Negative samples are selected by first identifying all classes in the subset, excluding the anchor’s class, and then sampling two different classes, selecting one instance from each. This class-aware sampling strategy guarantees semantically meaningful quadruplets, enabling the learning of discriminative and robust feature representations, particularly in scenarios with class imbalance or non-IID distributions, which are common in federated learning.
$\mathcal{D}=\left\{\left(x_i, y^i\right)\right\}_{i=1}^N \text { be the dataset with inputs } x_i \in \mathbb{R}^d \text { and labels } y^i \in\{0,1, \ldots, K-1\} .$
\begin{itemize}
    \item $x^a \sim \mathcal{D}_k$
    \item $x^p \sim \mathcal{D}_k \backslash\left\{x^a\right\}$
    \item $x^{n1} \sim \mathcal{D}_{k^{\prime}} \quad$ where $k^{\prime} \in\{0, \ldots, K-1\} \backslash\{k\}$
    \item  $x^{n2} \sim \mathcal{D}_{k^{\prime \prime}} \quad$ where $k^{\prime \prime} \in\{0, \ldots, K-1\} \backslash\left\{k, k^{\prime}\right\}$
\end{itemize}

This sampling approach facilitates training with the quadruplet loss, which enforces the following constraints in the embedding space:
\begin{itemize}
    \item Positive pairs $(x^a,x^p)$ are pulled close together (minimazing intra-class variance),
    \item Anchor $x^a$ is pushed away from two distinct negatives $(x^{n1},x^{n2})$ and enhancing robustness,
    \item Selecting two distinct negatives encourages higher inter-class variance.
\end{itemize}
   
Such structured sampling is particularly beneficial in metric learning and federated learning settings, where it improves generalisation and discriminative ability under data heterogeneity.

As illustrated in Figure \ref{fig:method}, our local loss comprises two parts. The first part is a typical supervised loss (e.g., cross-entropy loss) denoted as $\ell_{ce}$. The second part is our reformulated quadruplet loss denoted as $\ell_{quad*}$. We define our loss as
\begin{equation}
\ell=\ell_{\text {ce }}\left(w_i^t ;(x^a, y^a)\right)+\beta \ell_{\text {quad* }}\left(wi^t ; (x^a, x^p, x^{n1}, x^{n2})\right)
\end{equation}
where $\beta$ is a hyperparameter to maintain the impact of
quadruplet loss. Our local objective is to minimise
\begin{equation}
\begin{aligned}
\min_{w_i^t} \; \mathbb{E}_{(x^a, y^a) \sim D^i} \Big[ 
& \; \ell_{\text{ce}}(w_i^t ; (x^a, y^a)) \\
& + \beta \, \ell_{\text{quad*}}(w_i^t ; (x^a, x^p, x^{n1}, x^{n2}))
\Big]
\end{aligned}
\end{equation}
where $m_1$ and $m_2$ are the values of margins in the two terms and $y^a$  refers to the class label of image $x^a$, $z_a$ is representation. The margins define the minimum desired difference between the distance from the anchor to the negatives. Our proposed loss is defined as;
\begin{equation}
\begin{aligned}
\ell_{\text {quad* }}= &  \left[\left\|f\left(x^a\right)-f\left(x^p\right)\right\|_2-\left\|f\left(x^a\right)-f\left(x^{n1}\right)\right\|_2+m_1\right]_+ \\
& + \left[\left\|f\left(x^a\right)-f\left(x^p\right)\right\|_2-\left\|f\left(x^a\right)-f\left(x^{n2}\right)\right\|_2+m_2\right]_+
\end{aligned}
\end{equation}
where $[x]_+ = max(0, x)$, $f(\cdot)$ is a function to extract embeddings. We choose two distinct margin values in our loss to avoid enforcing the same separation radius for both negative samples. This allows more flexibility in the embedding space, preventing all negative instances from being pushed away uniformly, and instead adapting the separation based on their semantic or class-wise dissimilarity. 

The overall federated learning algorithm is shown in Algorithm \ref{alg:alg1} that represents the FedQuad design. FedQuad remains applicable when only a small subset of clients participates in each federated learning round.  We follow the standard FedAvg approach for model aggregation; each client maintains a local model, which is synchronised with the global model regularly and updated with the local models from the clients participating in that round.
\begin{algorithm}
\caption{FedQuad Framework}
\label{alg:alg1}
\KwIn{Dataset $D$, Number of communication rounds $T$, number of clients $N$, local epochs $E$, $B$ batch size, hyperparameters $\beta$, $m_1$, $m_2$}
\KwOut{Global model $w^T$}

\textbf{Server} \\
Initialize global model $w^0$ \\
\For{$t = 0, 1, \dots, T-1$}{
    \For{$i = 1$ \KwTo $N$}{
        Get global model $w^t$ to update client $C_i$ \\
        $w_i^t \leftarrow$ \textsc{\textbf{trainClient}}$(i, w^t)$
    }
    $w^{t+1} \leftarrow \sum_{k=1}^{N} \frac{|D_i|}{|D|} w_k^t$
}
\Return $w^T$

\vspace{1em}
\textbf{Function} \textsc{\textbf{trainClient}}$(i, w^t)$:
\BlankLine
$w_i^t \leftarrow w^t$ \\
\For{$e = 1$ \KwTo $E$}{
    \For{each batch $b = \{x^a_i, x^p_i, x^{n1}_i, x^{n2}_i, y^a_i\}_{i < B}$ from $\mathcal{D}_i$}{
 
        $z_a \leftarrow f_{w_i^t}(x^a)$ \\
        $z_{\text{p}} \leftarrow f_{w_i^t}(x^p)$ \\
        $z_{\text{n1}} \leftarrow f_{w_i^t}(x^{n1})$ \\
        $z_{\text{n2}} \leftarrow f_{w_i^t}(x^{n2})$ \\
        $\ell_{\text{quad*}} \leftarrow \left[d\left(z_a, z_p\right)^2-d\left(z_a, z_{n1}\right)^2+ m_1\right]_{+} + \left[ d\left(z_a, z_p\right)^2-d\left(z_a, z_{n2}\right)^2+ m_2\right]_{+} $ \\
        $\ell_{\text{ce}} \leftarrow \text{CrossEntropyLoss}(F_{w_i^t}(x^a), y^a)$ \\
        $\ell \leftarrow \ell_{\text{ce}} + \beta \cdot \ell_{\text{quad*}}$ \\
        $w_i^t \leftarrow w_i^t - \eta \cdot \nabla \ell$
    }
}
\Return $w_i^t$
\end{algorithm}
%Table \ref{tab:c10all},\ref{tab:c100all} our version of QuadrupletFL and FedQuad, consistently outperforms the standard supervised federated metric learning baseline. Although the modified loss function introduces twice as two negative pairs compared to the traditional triplet loss, it achieves superior performance by effectively leveraging a richer set of negative relationships. While triplet loss focuses on a single positive-negative pair at a time, incorporating multiple negative samples better facilitates the maximisation of inter-class variance, leading to more discriminative representations.

\section{Experiment}
\label{sec:4}
We evaluate our proposed method against standard supervised metric learning approaches, including Triplet loss, Supervised Contrastive loss, and Quadruplet loss. To ensure a fair comparison, all baselines are implemented and evaluated within the same federated learning setting and reproducibility rules (e.g. seeds). Additionally, we include a fully supervised baseline trained on the entire dataset. Experiments are conducted on the CIFAR-10, CIFAR-100, and Tiny-ImageNet datasets \cite{krizhevsky2009learning}, \cite{tinyimage}.
\subsection{Experimental Setup}
Our model employs a convolutional neural network (CNN) backbone followed by a fully connected layer to produce embeddings of a specified dimension. The CNN backbone consists of three convolutional blocks. Each block contains a 2D convolutional layer (kernel size 3$\times$3, stride 1, padding 1), followed by batch normalization and a ReLU activation. The first two blocks conclude with a 2$\times$2 max pooling layer to reduce spatial dimensions, while the final block uses an adaptive average pooling layer to produce a fixed-size output of 1$\times$1. The output of the convolutional backbone is flattened and passed through a fully connected layer that maps the 256-dimensional feature vector to the desired embedding dimension, which is 128 in our experiments. A softmax layer is included solely for cross-entropy loss measurement. The code for our experiments is publicly available to ensure reproducibility\footnote{https://anonymous.4open.science/r/FedQuad-55C8/README.md}.
We use the Adam optimizer with a learning rate of 0.001 for all approaches. Weight decay is set to $10^{-5}$, and momentum is fixed at 0.9. The batch size is 128. For all federated learning approaches, the number of local epochs is set to 5 unless otherwise specified. The number of communication rounds is 20 for each dataset, as additional rounds yield negligible improvement in accuracy.

Following prior work \cite{fedavg, li2021model}, we employ a Dirichlet distribution to generate non-IID data partitions among clients, with concentration parameters $\alpha=0.5$ and $\alpha=0.3$ (lower $\alpha$ corresponds to a more skewed distribution). This partitioning results in some clients having few or even no samples for certain classes. We evaluate scenarios with 10, 50, and 200 clients, considering diverse data distributions, as reported in Table \ref{tab:cifar10} (CIFAR-10), Table \ref{tab:cifar100} (CIFAR-100), and Table \ref{tab:tiny} (TinyImageNet). For our reformulated quadruplet loss, the best hyperparameters are $\beta=0.5$, with margin values $m_1=1.0$ and $m_2=0.5$, as shown in the ablation study (Table \ref{tab:abl}).

\section{Results}
\label{sec:5}
We evaluate FedQuad against several metric learning-based federated learning baselines under varying numbers of clients and data distribution settings, using CIFAR-10, CIFAR-100 and TinyImageNet. 

As a baseline comparison, we first evaluate the same experimental settings using centralised supervised metric learning methods to better understand the impact of federated learning on metric learning-based approaches under comparable conditions. The results in Table \ref{tab:cifar10_100_tiny} show that the fully supervised setting achieves the best image classification accuracy on CIFAR-10 and TinyImageNet. However, under IID conditions on CIFAR-100, the federated approach FedQuad outperforms its supervised counterpart, demonstrating its effectiveness in more complex, high-class-cardinality scenarios.

As reported in Table \ref{tab:cifar10_100_tiny}, FedQuad consistently outperforms all baselines across different levels of data heterogeneity. Table \ref{tab:cifar10_3}, \ref{tab:cifar100_3}, and \ref{tab:tiny3} further demonstrate that our method maintains high performance even under highly non-IID distributions with a large number of clients. These results indicate that distance-aware optimisation of data representations improves feature separability, resulting in more compact intra-class clusters and clearer inter-class boundaries. Tables \ref{tab:cifar10_3}, \ref{tab:cifar100_3}, and \ref{tab:tiny3} show that increasing data heterogeneity, combined with a larger number of clients, makes representation learning more challenging. This effect is particularly pronounced in datasets with a higher number of classes. Under high heterogeneity, learning discriminative representations becomes more difficult for CIFAR-100 and TinyImageNet, which contain 100 and 200 classes, respectively, compared to CIFAR-10, which has only 10 classes. 

FedQuad explicitly maximises inter-class variance within each client’s representation space, enhancing feature discrimination. This is particularly advantageous when evaluating the global model on test data that spans all clients. Even when certain clients lack samples for specific classes, FedQuad enables the global model to learn robust and generalizable representations.

\begin{table*}[htbp]
\centering
\footnotesize
\renewcommand{\arraystretch}{1.15}

\begin{subtable}[t]{0.47\textwidth}
\centering
\begin{tabular}{lccc}
\toprule
\textbf{Method} & \textbf{IID} & $\boldsymbol{\alpha=0.5}$ & $\boldsymbol{\alpha=0.3}$ \\
\midrule
\textit{\color{gray}Supervised (Non-FL)} \\
Supervised  & \textbf{82.57} & -- & -- \\
SupCon      & 76.67 & -- & -- \\
Triplet     & 64.52 & -- & -- \\
Quadruplet  & 67.72 & -- & -- \\
\midrule
\textit{\color{gray}Supervised FL} \\
FedAvg       & 81.26  & 77.34 & 74.05\\
SupConFL     & 71.06  & 69.46 & 68.07\\
TripletFL    & 59.67  & 60.77 & 40.42\\
QuadrupletFL & 62.79  & 64.18 & 47.27 \\
MOON         & 76.86  & 61.43 & 49.69 \\
\textbf{FedQuad} & \textbf{82.35}  & \textbf{80.83} & \textbf{80.13}\\
\bottomrule
\end{tabular}
\caption{CIFAR-10 results (10 clients).}
\label{tab:cifar10}
\end{subtable}
\hfill
\begin{subtable}[t]{0.47\textwidth}
\centering
\begin{tabular}{lccc}
\toprule
\textbf{Method} & \textbf{IID} & $\boldsymbol{\alpha=0.5}$ & $\boldsymbol{\alpha=0.3}$ \\
\midrule
\textit{\color{gray}Supervised (Non-FL)} \\
Supervised  & \textbf{50.96} & -- & -- \\
SupCon      & 37.33 & -- & -- \\
Triplet     & 39.43 & -- & -- \\
Quadruplet  & 40.33 & -- & -- \\
\midrule
\textit{\color{gray}Supervised FL} \\
FedAvg       & 51.33 & 47.56  & 44.32 \\
SupConFL     & 34.79  & 36.21 & 36.72\\
TripletFL    & 36.29  & 35.26 & 33.01\\
QuadrupletFL & 39.49  & 35.96 & 33.92 \\
MOON         & 26.32  & 25.73 & 26.10 \\
\textbf{FedQuad} & \textbf{51.27}  & \textbf{50.64} & \textbf{48.55} \\
\bottomrule
\end{tabular}
\caption{CIFAR-100 results (10 clients).}
\label{tab:cifar100}
\end{subtable}

\vspace{0.5cm} % space between rows

% --- Second row (1 centred subtable) ---
\begin{subtable}[t]{0.47\textwidth}
\centering
\begin{tabular}{lccc}
\toprule
\textbf{Method} & \textbf{IID} & $\boldsymbol{\alpha=0.5}$ & $\boldsymbol{\alpha=0.3}$ \\
\midrule
\textit{\color{gray}Supervised (Non-FL)} \\
Supervised  & \textbf{70.83} & -- & -- \\
SupCon      & 17.92 & -- & -- \\
Triplet     & 30.08 & -- & -- \\
Quadruplet  & 33.56 & -- & -- \\
\midrule
\textit{\color{gray}Supervised FL} \\
FedAvg       & 36.12  & 32.52 & 29.56 \\
SupConFL      & 15.70    & 15.42 & 15.52\\
TripletFL     & 22.52 & 21.22 & 21.20  \\
QuadrupletFL & 35.78   &  23.26  & 21.74\\
MOON         & 26.82 & 24.54  & 21.88  \\
\textbf{FedQuad} & \textbf{35.82}  & \textbf{35.32} & \textbf{35.06} \\
\bottomrule
\end{tabular}
\caption{Tiny-ImageNet results (10 clients).}
\label{tab:tiny}
\end{subtable}

\caption{Test accuracy (\%) comparison across supervised and federated learning variants on CIFAR-10, CIFAR-100, and Tiny-ImageNet under varying data heterogeneity ($\alpha$).}
\label{tab:cifar10_100_tiny}
\end{table*}

Tables \ref{tab:cifar10_3}, \ref{tab:cifar100_3} and \ref{tab:tiny3} illustrate that an increasing number of clients can exacerbate data sparsity and amplify distributional diversity, which in turn negatively affects the performance of many existing methods. This phenomenon is particularly pronounced on CIFAR-100, which contains 100 classes with only 500 samples per class. When data is partitioned among 200 clients, the number of samples per class per client becomes extremely limited, making it challenging to learn robust feature representations.

Despite these difficulties, FedQuad effectively mitigates both data imbalance and sparsity under non-IID data distributions. In contrast, MOON exhibits substantially degraded performance under similar conditions, highlighting the limitations of global-model-regularization–based contrastive learning in scenarios with many clients and highly heterogeneous data. These results indicate that MOON struggles to generalise and fails to learn robust representations under severe data heterogeneity.

\begin{table}[ht]
\centering
\scriptsize
\renewcommand{\arraystretch}{1.2}
\setlength{\tabcolsep}{4pt}
\begin{tabular}{l|ccc|ccc|ccc}
\toprule
\textbf{Method} &
\multicolumn{3}{c|}{IID} &
\multicolumn{3}{c|}{$\boldsymbol{\alpha=0.5}$} &
\multicolumn{3}{c}{\textbf{$\boldsymbol{\alpha=0.3}$}} \\
\cmidrule(lr){2-4}\cmidrule(lr){5-7}\cmidrule(lr){8-10}
 & 10C & 50C & 200C & 10C & 50C & 200C & 10C & 50C & 200C \\
\midrule
FedAvg        & 79.86{\tiny$\pm$0.28} & 70.53{\tiny$\pm$0.09} & 61.46{\tiny$\pm$0.44} & 76.85{\tiny$\pm$0.86} & 68.61{\tiny$\pm$0.46} & 59.82{\tiny$\pm$0.21} & 74.03{\tiny$\pm$1.37} & 67.6{\tiny$\pm$0.67} & 58.57{\tiny$\pm$0.76} \\
SupConFL      & 71.55{\tiny$\pm$0.77} & 50.27{\tiny$\pm$0.08} & 46.92{\tiny$\pm$0.51} & 68.88{\tiny$\pm$3.33} & 49.75{\tiny$\pm$0.05} & 46.38{\tiny$\pm$0.41} & 69.07{\tiny$\pm$1.32} & 49.53{\tiny$\pm$0.21} & 46.29{\tiny$\pm$0.03} \\
TripletFL     & 74.25{\tiny$\pm$0.40} & 61.23{\tiny$\pm$0.10} & 54.86{\tiny$\pm$0.62} & 64.60{\tiny$\pm$3.68} & 60.48{\tiny$\pm$0.58} & 54.25{\tiny$\pm$0.30} & 55.53{\tiny$\pm$3.10} & 57.97{\tiny$\pm$0.25} & 54.46{\tiny$\pm$0.23} \\
QuadrupletFL  & 76.77{\tiny$\pm$0.28} & 72.32{\tiny$\pm$0.17} & 61.66{\tiny$\pm$0.65} & 70.30{\tiny$\pm$0.14} & 59.45{\tiny$\pm$0.43} & \textbf{61.12}{\tiny$\pm$2.10} & 58.79{\tiny$\pm$2.42} & \textbf{68.68}{\tiny$\pm$0.62} & 58.76{\tiny$\pm$0.01} \\
MOON          & 79.03{\tiny$\pm$0.11} & 71.02{\tiny$\pm$0.33} & 60.51{\tiny$\pm$0.57} & 69.23{\tiny$\pm$2.63} & 69.36{\tiny$\pm$0.22} & 59.94{\tiny$\pm$0.09} & 64.19{\tiny$\pm$2.35} & 68.04{\tiny$\pm$0.46} & 59.08{\tiny$\pm$0.32} \\
\textbf{FedQuad} & \textbf{82.37}{\tiny$\pm$0.25} & \textbf{72.37}{\tiny$\pm$0.17} & \textbf{63.02}{\tiny$\pm$0.01} & \textbf{80.76}{\tiny$\pm$0.22} & \textbf{69.80}{\tiny$\pm$0.19} & {59.86}{\tiny$\pm$0.21} & \textbf{79.45}{\tiny$\pm$0.67} & {68.03}{\tiny$\pm$0.79} & \textbf{59.46}{\tiny$\pm$0.21} \\
\bottomrule
\end{tabular}
\caption{Test accuracy (\%) of different methods on CIFAR-10 under varying data heterogeneity and client numbers. The bold values indicate the best results for each setting.}
\label{tab:cifar10_3}
\end{table}

\begin{table}[ht]
\centering
\scriptsize
\renewcommand{\arraystretch}{1.2}
\setlength{\tabcolsep}{4pt}
\begin{tabular}{l|ccc|ccc|ccc}
\toprule
\textbf{Method} &
\multicolumn{3}{c|}{IID} &
\multicolumn{3}{c|}{$\boldsymbol{\alpha=0.5}$} &
\multicolumn{3}{c}{\textbf{$\boldsymbol{\alpha=0.3}$}} \\
\cmidrule(lr){2-4}\cmidrule(lr){5-7}\cmidrule(lr){8-10}
 & 10C & 50C & 200C & 10C & 50C & 200C & 10C & 50C & 200C \\
\midrule
FedAvg        & 51.52{\tiny$\pm$0.15} & 35.23{\tiny$\pm$0.17} & 24.86{\tiny$\pm$0.84}   & 47.75{\tiny$\pm$0.31} & 36.28{\tiny$\pm$0.13} & 24.33{\tiny$\pm$0.13} & 44.95{\tiny$\pm$0.44} & 35.23{\tiny$\pm$0.17} & 23.84{\tiny$\pm$0.21}\\
SupConFL    &  35.05{\tiny$\pm$0.21} & 27.28{\tiny$\pm$0.56} & 20.22{\tiny$\pm$0.21} & 36.00{\tiny$\pm$0.21} & 27.01{\tiny$\pm$0.58} & 19.34{\tiny$\pm$0.11} & 36.41{\tiny$\pm$0.22} & 26.87{\tiny$\pm$0.32} & 19.01{\tiny$\pm$0.18} \\
TripletFL    & 36.56{\tiny$\pm$0.19} & 27.56{\tiny$\pm$0.12} & 23.83{\tiny$\pm$0.29} & 35.10{\tiny$\pm$0.47} & 30.89{\tiny$\pm$4.56} & 23.63{\tiny$\pm$0.74} & 33.96{\tiny$\pm$0.96} & 28.42{\tiny$\pm$0.15} & 24.26{\tiny$\pm$0.15}  \\
QuadrupletFL  & 37.85{\tiny$\pm$0.45} & 29.67{\tiny$\pm$0.41} & 25.83{\tiny$\pm$0.47}  & 34.69{\tiny$\pm$0.89} & 29.64{\tiny$\pm$0.09} & 25.53{\tiny$\pm$0.38} & 32.21{\tiny$\pm$1.21} & 29.93{\tiny$\pm$0.24} & 25.57{\tiny$\pm$0.63} \\
MOON          & 37.09{\tiny$\pm$0.61} & 23.72{\tiny$\pm$0.02} & 14.69{\tiny$\pm$0.32} & 32.80{\tiny$\pm$0.35} & 21.46{\tiny$\pm$0.16} & 13.63{\tiny$\pm$0.36} & 31.43{\tiny$\pm$1.72} & 20.09{\tiny$\pm$0.46} & 13.06{\tiny$\pm$0.02} \\
\textbf{FedQuad} &  \textbf{51.49}{\tiny$\pm$0.16} & \textbf{36.59}{\tiny$\pm$0.41} & \textbf{26.07}{\tiny$\pm$0.17} & \textbf{50.77}{\tiny$\pm$0.24} & \textbf{31.55}{\tiny$\pm$5.32} & \textbf{26.95}{\tiny$\pm$0.31} & \textbf{50.56}{\tiny$\pm$0.04} & \textbf{34.96}{\tiny$\pm$0.41} & \textbf{26.90}{\tiny$\pm$0.16} \\

\bottomrule
\end{tabular}
\caption{Test accuracy (\%) of different methods on CIFAR-100 under varying data heterogeneity and client numbers. The bold values indicate the best results for each setting.}
\label{tab:cifar100_3}
\end{table}

Figures \ref{fig:ratioc10}, \ref{fig:ratioc100}, and \ref{fig:ratiotiny} illustrate the ratio of inter-class to intra-class distances, along with the corresponding average inter-class distance values for datasets. These figures show that the reformulated quadruplet loss captures robust embeddings by effectively modelling the distances between positive and negative pairs. Across all settings, FedQuad achieves the lowest intra-class variance, indicating strong feature separation within each class.

In contrast, TripletFL exhibits a higher inter-class to intra-class ratio; however, this is accompanied by relatively larger intra-class distances. This suggests that while inter-class variance is decreased, intra-class variance is not sufficiently preserved. Overall, these results indicate that incorporating multiple negative relationships, as in FedQuad, leads to more discriminative and well-structured representations in the global feature space.

In the CIFAR-10 experiments, across various data distribution settings, SupConFL demonstrates relatively weak performance, exhibiting poor discriminative ability between similar and dissimilar feature representations. A higher ratio indicates stronger class separability, which is observed in TripletFL with around 1.5. However, TripletFL also exhibits the highest average intra-class distance, suggesting that although maximising inter-class variance is substantial, samples within the same class remain dispersed, leading to suboptimal intra-class variance.

\begin{table}[ht]
\centering
\scriptsize
\renewcommand{\arraystretch}{1.2}
\setlength{\tabcolsep}{4pt}
\begin{tabular}{l|ccc|ccc|ccc}
\toprule
\textbf{Method} &
\multicolumn{3}{c|}{IID} &
\multicolumn{3}{c|}{$\boldsymbol{\alpha=0.5}$} &
\multicolumn{3}{c}{\textbf{$\boldsymbol{\alpha=0.3}$}} \\
\cmidrule(lr){2-4}\cmidrule(lr){5-7}\cmidrule(lr){8-10}
 & 10C & 50C & 200C & 10C & 50C & 200C & 10C & 50C & 200C \\
\midrule
FedAvg        & 35.62{\tiny$\pm 0.45$} & 24.77{\tiny$\pm 0.08$} & 15.82{\tiny$\pm 0.14$} & 31.99{\tiny$\pm 0.37$} & \textbf{23.33{\tiny$\pm 0.16$}} & \textbf{15.25{\tiny$\pm 0.40$}} & 29.36{\tiny$\pm 0.16$} & \textbf{22.65{\tiny$\pm 0.21$}} & 14.44{\tiny$\pm 0.28$} \\
SupConFL      & 15.40{\tiny$\pm 0.35$} & 13.15{\tiny$\pm 0.16$} & 13.42{\tiny$\pm 1.42$} & 15.63{\tiny$\pm 0.21$} & 13.20{\tiny$\pm 0.02$} & 11.98{\tiny$\pm 0.02$} & 15.59{\tiny$\pm 0.17$} & 13.32{\tiny$\pm 0.18$} & 12.14{\tiny$\pm 0.11$} \\
TripletFL     & 22.87{\tiny$\pm 0.56$} & 17.06{\tiny$\pm 0.19$} & 14.63{\tiny$\pm 0.44$} & 22.09{\tiny$\pm 0.67$} & 17.38{\tiny$\pm 0.47$} & 14.85{\tiny$\pm 0.27$} & 21.15{\tiny$\pm 0.40$} & 17.22{\tiny$\pm 0.51$} & \textbf{14.62{\tiny$\pm 0.36$}} \\
QuadrupletFL  & 35.33{\tiny$\pm 0.57$} & 24.46{\tiny$\pm 0.12$} & 15.83{\tiny$\pm 0.33$} & 23.17{\tiny$\pm 0.29$} & 23.16{\tiny$\pm 0.33$} & 13.81{\tiny$\pm 0.17$} & 21.64{\tiny$\pm0.00$} & 22.62{\tiny$\pm 0.17$} & 13.34{\tiny$\pm 0.35$} \\
MOON          & 26.84{\tiny$\pm0.20$} & \textbf{22.67{\tiny$\pm$1.53}} & \textbf{21.70{\tiny$\pm0.26$}} & 16.69{\tiny$\pm0.13$} &  14.17{\tiny$\pm0.47$} & 12.61{\tiny$\pm0.19$} & 11.05{\tiny$\pm0.11$} & 9.94{\tiny$\pm0.22$} & 11.05{\tiny$\pm0.11$} \\
\textbf{FedQuad} & \textbf{35.92{\tiny$\pm 0.18$}} & 16.90{\tiny$\pm 0.12$} & 12.98{\tiny$\pm2.82$} & \textbf{35.32{\tiny$\pm 0.23$}} & 13.38{\tiny$\pm 0.48$} & 10.53{\tiny$\pm 0.03$} & \textbf{34.99{\tiny$\pm 0.37$}} &12.76{\tiny$\pm 0.19$} & 10.03{\tiny$\pm0.20$} \\
\bottomrule
\end{tabular}
\caption{Test accuracy (\%) of different methods on Tiny-ImageNet under varying data heterogeneity and client numbers. The bold values indicate the best results for each setting.}
\label{tab:tiny3}
\end{table}
In contrast, FedQuad achieves the lowest intra-class distance, indicating that features of similar samples are tightly clustered. This balance between minimising intra-class variance and maintaining sufficient inter-class separation highlights the superior representational consistency of FedQuad compared to other methods. 

The CIFAR-100 figure \ref{fig:ratioc100}, reveals that FedAvg exhibits the highest intra-class distance, indicating its limited ability to handle highly diverse datasets, particularly when the number of classes increases tenfold compared to CIFAR-10. This suggests that FedAvg struggles to achieve effective feature separation under such complexity. However, when examining the pairwise distance distributions, FedAvg demonstrates slightly improved discrimination across certain class pairs. 

Figure \ref{fig:ratiotiny} illustrates that despite training, the learned representations suffer from representational collapse and poor embedding quality, characterized by high intra-class variance and insufficient class separation. This highlights the vulnerability of feature embeddings to the global aggregation process in federated learning. Consequently, acquiring a sufficient number of quadruplet sets becomes crucial for enforcing effective distance-based separation within the feature space.

FedQuad consistently achieves a balanced inter-class to intra-class distance ratio across all three data distribution settings. Its notably lower intra-class distance highlights its superior capacity to cluster similar samples more compactly while maintaining clear separation between distinct classes.

%%%%%%%%%%%%%%%%%%%%%%%%%%%%%%%%%%%%%%%%%%%%%%%%%
\begin{figure}[htbp]
\raggedright
\setlength{\tabcolsep}{6.5pt} % adjust spacing between columns
\renewcommand{\arraystretch}{1.0}
\begin{tabular}{ccc}

% ---------- (a) ----------
\begin{tikzpicture}
\begin{axis}[
    title={\footnotesize IID},
    width=5cm,
    height=6cm,
    ymin=0, ymax=2.5,
    xlabel={\footnotesize Methods},
    xticklabel style={font=\scriptsize, rotate=45, anchor=east},
    xlabel style={at={(axis description cs:0.5,-0.18)}, anchor=north},
    symbolic x coords={FedAvg, SupConFL, TripletFL, QuadrupletFL, MOON, FedQuad},
    xtick=data,
    bar width=10pt,
    ymajorgrids=true,
    grid style={dashed,gray!40},
    enlarge x limits=0.2,
    extra description/.append code={
        \node[rotate=-90, anchor=south, font=\scriptsize]
        at (rel axis cs:1.15,0.5) {Intra-Class Distance};
    },
    extra description/.append code={
    \node[rotate=90, anchor=south, font=\scriptsize]
    at (rel axis cs:-0.14,0.5) {Inter/Intra Class Ratio};
},
]
\addplot[ybar, fill=cyan!20!white, draw=cyan!60!black,
    error bars/.cd, y dir=both, y explicit]
coordinates {
    (FedAvg,1.57) +- (0.27,0.27)
    (SupConFL,1.14) +- (0.6,0.6)
    (TripletFL,1.63) +- (0.4,0.4)
    (QuadrupletFL,1.47) +- (0.07,0.07)
    (MOON,1.32) +- (0.24,0.24)
    (FedQuad,1.49) +- (0.11,0.11)
};
\end{axis}
\begin{axis}[
    width=5cm, height=6cm,
    ymin=0, ymax=0.8,
    axis x line=none,
    axis y line*=right,
    yticklabel style={font=\scriptsize},
    symbolic x coords={FedAvg, SupConFL, TripletFL, QuadrupletFL, MOON, FedQuad},
    xtick=data,
    enlarge x limits=0.2,
     legend style={
        at={(0.96,0.94)}, anchor=north east,
        font=\tiny, draw=none, fill=white, fill opacity=1.0
    }
]
\addplot[thick, mark=*, mark options={fill=blue!60!black}, draw=blue!60!black]
coordinates {
    (FedAvg,0.35) (SupConFL,0.32) (TripletFL,0.64)
    (QuadrupletFL,0.13) (MOON,0.27) (FedQuad,0.15)
};
\addlegendimage{thick, mark=*, draw=none, mark options={fill=blue!60!black}}
\legend{\scriptsize Intra}
\end{axis}
\end{tikzpicture}
&
% ---------- (b) ----------
\begin{tikzpicture}
\begin{axis}[
    title={\footnotesize $\alpha=0.5$},
    width=5cm,
    height=6cm,
    ymin=0, ymax=2.5,
    xlabel={\footnotesize Methods},
    symbolic x coords={FedAvg, SupConFL, TripletFL, QuadrupletFL, MOON, FedQuad},
    xtick=data,
    xticklabel style={font=\scriptsize, rotate=45, anchor=east},
    xlabel style={at={(axis description cs:0.5,-0.18)}, anchor=north},
    bar width=10pt,
    ymajorgrids=true,
    grid style={dashed,gray!40},
    enlarge x limits=0.2,
        extra description/.append code={
        \node[rotate=-90, anchor=south, font=\scriptsize]
        at (rel axis cs:1.15,0.5) {Intra-Class Distance};
    },
    extra description/.append code={
    \node[rotate=90, anchor=south, font=\scriptsize]
    at (rel axis cs:-0.14,0.5) {Inter/Intra Class Ratio};
},
]
\addplot[ybar, fill=cyan!20!white, draw=cyan!60!black,
    error bars/.cd, y dir=both, y explicit]
coordinates {
    (FedAvg,1.39) +- (0.19,0.19)
    (SupConFL,1.13) +- (0.39,0.39)
    (TripletFL,1.60) +- (0.07,0.07)
    (QuadrupletFL,1.47) +- (0.45,0.45)
    (MOON,1.28) +- (0.54,0.54)
    (FedQuad,1.47) +- (0.19,0.19)
};
\end{axis}
\begin{axis}[
    width=5cm, height=6cm,
    ymin=0, ymax=0.8,
    axis x line=none,
    axis y line*=right,
    yticklabel style={font=\scriptsize},
    symbolic x coords={FedAvg, SupConFL, TripletFL, QuadrupletFL, MOON, FedQuad},
    xtick=data,
    enlarge x limits=0.2,
     legend style={
        at={(0.96,0.94)}, anchor=north east,
        font=\tiny, draw=none, fill=white, fill opacity=1.0
    }
]
\addplot[thick, mark=*, mark options={fill=blue!60!black}, draw=blue!60!black]
coordinates {
    (FedAvg,0.38) (SupConFL,0.39) (TripletFL,0.61)
    (QuadrupletFL,0.11) (MOON,0.11) (FedQuad,0.15)
};
\addlegendimage{thick, mark=*, draw=none, mark options={fill=blue!60!black}}
\legend{\scriptsize Intra}
\end{axis}
\end{tikzpicture}
&
% ---------- (c) ----------
\begin{tikzpicture}
\begin{axis}[
    title={\footnotesize $\alpha=0.3$},
    width=5cm,
    height=6cm,
    ymin=0, ymax=2.5,
    xlabel={\footnotesize Methods},
    symbolic x coords={FedAvg, SupConFL, TripletFL, QuadrupletFL, MOON, FedQuad},
    xtick=data,
    xticklabel style={font=\scriptsize, rotate=45, anchor=east},
    xlabel style={at={(axis description cs:0.5,-0.18)}, anchor=north},
    bar width=10pt,
    ymajorgrids=true,
    grid style={dashed,gray!40},
    enlarge x limits=0.2,
        extra description/.append code={
        \node[rotate=-90, anchor=south, font=\scriptsize]
        at (rel axis cs:1.15,0.5) {Intra-Class Distance};
    },
    extra description/.append code={
    \node[rotate=90, anchor=south, font=\scriptsize]
    at (rel axis cs:-0.14,0.5) {Inter/Intra Class Ratio};
},
]
\addplot[ybar, fill=cyan!20!white, draw=cyan!60!black,
    error bars/.cd, y dir=both, y explicit]
coordinates {
    (FedAvg,1.33) +- (0.45,0.45)
    (SupConFL,1.13) +- (0.45,0.45)
    (TripletFL,1.58) +- (0.2,0.2)
    (QuadrupletFL,1.46) +- (0.54,0.54)
    (MOON,1.26) +- (0.09,0.09)
    (FedQuad,1.45) +- (0.24,0.24)
};
\end{axis}

\begin{axis}[
    width=5cm, height=6cm,
    ymin=0, ymax=0.8,
    axis x line=none,
    axis y line*=right,
    yticklabel style={font=\scriptsize},
    symbolic x coords={FedAvg, SupConFL, TripletFL, QuadrupletFL, MOON, FedQuad},
    xtick=data,
    enlarge x limits=0.2,
     legend style={
        at={(0.96,0.94)}, anchor=north east,
        font=\tiny, draw=none, fill=white, fill opacity=1.0
    }
]
\addplot[thick, mark=*, mark options={fill=blue!60!black}, draw=blue!60!black]
coordinates {
    (FedAvg,0.29) (SupConFL,0.42) (TripletFL,0.57)
    (QuadrupletFL,0.098) (MOON,0.048) (FedQuad,0.096)
};
\addlegendimage{thick, mark=*, draw=none, mark options={fill=blue!60!black}}
\legend{\scriptsize Intra}
\end{axis}
\end{tikzpicture}

\end{tabular}
\caption{
Inter/Intra-class ratio (↑ better) across federated learning methods on CIFAR-10  
(200 clients). Error bars indicate standard deviation over runs.
}
\label{fig:ratioc10}
\end{figure}
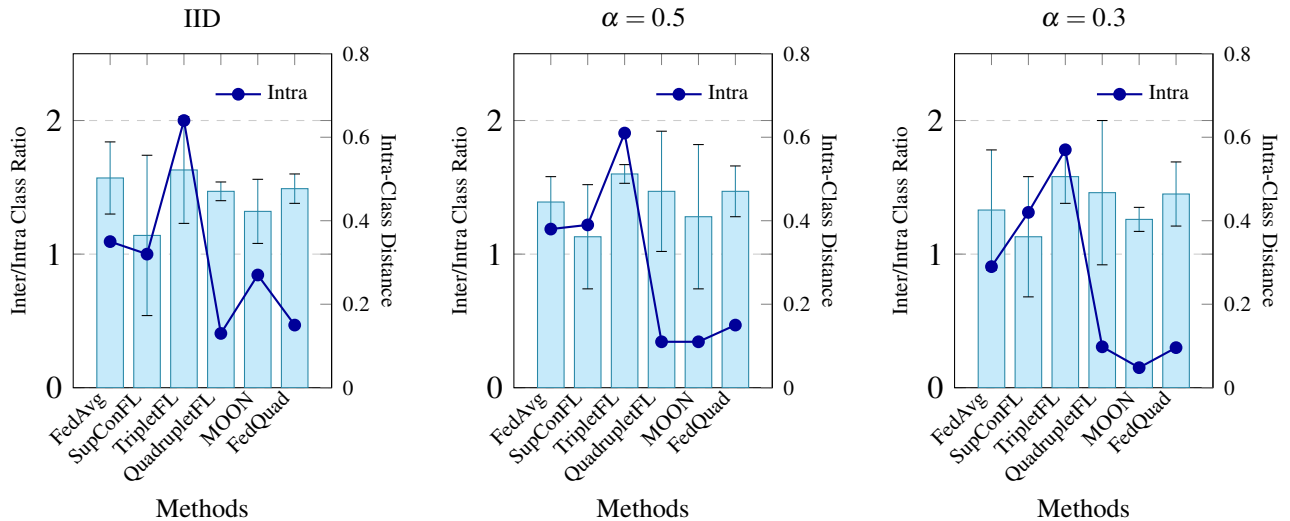

%%%%%%%%%%%%%%%%%%%%%%%%%%%%%%%%%%%%%%%%%%%%%%%%%
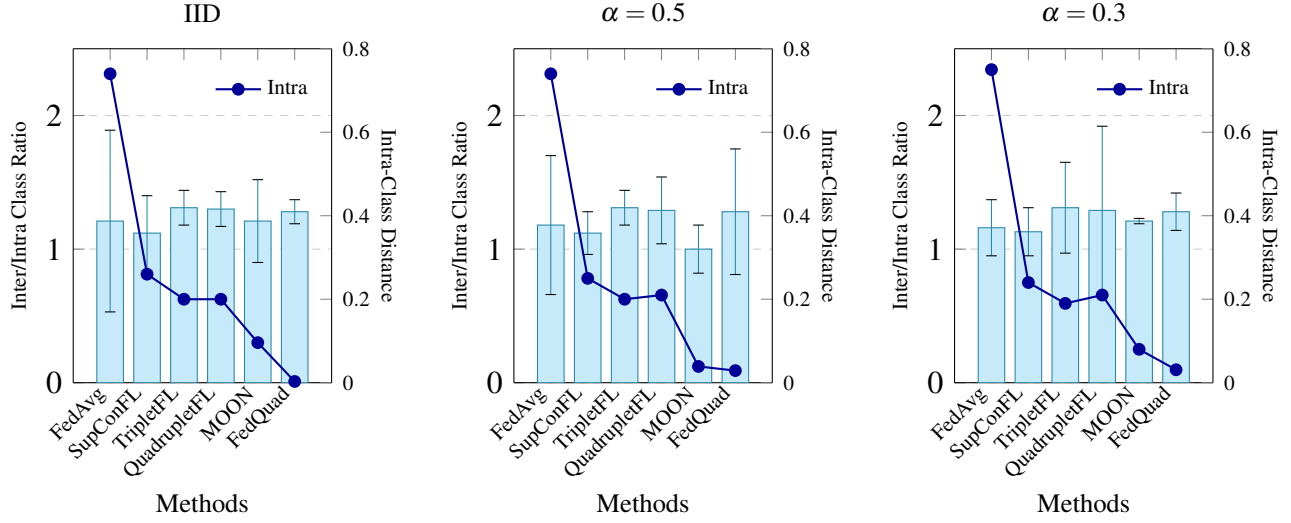
\begin{figure}[htbp]
\raggedright
\setlength{\tabcolsep}{6.5pt} % adjust spacing between columns
\renewcommand{\arraystretch}{1.0}
\begin{tabular}{ccc}

% ---------- (a) ----------
\begin{tikzpicture}[baseline]
\begin{axis}[
    title={\footnotesize IID},
    width=5cm, height=6cm,
    ymin=0, ymax=2.5,
    axis y line* = left,
    %ylabel={\footnotesize Inter/Intra Class Ratio},
    %yticklabel style={font=\scriptsize},
    %every axis y label/.append style={xshift=-0.7ex},
    xlabel={\footnotesize Methods},
    xticklabel style={font=\scriptsize, rotate=45, anchor=east},
    xlabel style={at={(axis description cs:0.5,-0.18)}, anchor=north},
    symbolic x coords={FedAvg, SupConFL, TripletFL, QuadrupletFL, MOON, FedQuad},
    xtick=data,
    bar width=10pt,
    ymajorgrids=true,
    grid style={dashed,gray!40},
    enlarge x limits=0.2,
    % --- right-side title ---
    extra description/.append code={
        \node[rotate=-90, anchor=south, font=\scriptsize]
        at (rel axis cs:1.15,0.5) {Intra-Class Distance};
    },
    extra description/.append code={
    \node[rotate=90, anchor=south, font=\scriptsize]
    at (rel axis cs:-0.14,0.5) {Inter/Intra Class Ratio};
},
]
\addplot[ybar, fill=cyan!20!white, draw=cyan!60!black,
 error bars/.cd, y dir=both, y explicit]
coordinates {
    (FedAvg,1.21) +- (0.68,0.68)
    (SupConFL,1.12) +- (0.28,0.28)
    (TripletFL,1.31) +- (0.13,0.13)
    (QuadrupletFL,1.30) +- (0.13,0.13)
    (MOON,1.21) +- (0.31,0.31)
    (FedQuad,1.28) +- (0.09,0.09)
};
\end{axis}

% --- Line overlay ---
\begin{axis}[
    width=5cm, height=6cm,
    ymin=0, ymax=0.8,
    axis x line=none,
    axis y line*=right,
    yticklabel style={font=\scriptsize},
    symbolic x coords={FedAvg, SupConFL, TripletFL, QuadrupletFL, MOON, FedQuad},
    xtick=data,
    enlarge x limits=0.2,
    legend style={
        at={(0.96,0.94)}, anchor=north east,
        font=\tiny, draw=none, fill=white, fill opacity=1.0
    }
]
\addplot[thick, mark=*, mark options={fill=blue!60!black}, draw=blue!60!black]
coordinates {
    (FedAvg,0.74) (SupConFL,0.26) (TripletFL,0.20)
    (QuadrupletFL,0.20) (MOON,0.096) (FedQuad,0.003)
};
\addlegendimage{thick, mark=*, draw=none, mark options={fill=blue!60!black}}
\legend{\scriptsize Intra}
\end{axis}
\end{tikzpicture}
&
% ---------- (b) ----------
\begin{tikzpicture}[baseline]
\begin{axis}[
    title={\footnotesize $\alpha=0.5$},
    width=5cm, height=6cm,
    ymin=0, ymax=2.5,
    axis y line*=left,
    xlabel={\footnotesize Methods},
    xticklabel style={font=\scriptsize, rotate=45, anchor=east},
    xlabel style={at={(axis description cs:0.5,-0.18)}, anchor=north},
    symbolic x coords={FedAvg, SupConFL, TripletFL, QuadrupletFL, MOON, FedQuad},
    xtick=data,
    bar width=10pt,
    ymajorgrids=true,
    grid style={dashed,gray!40},
    enlarge x limits=0.2,
    extra description/.append code={
        \node[rotate=-90, anchor=south, font=\scriptsize]
        at (rel axis cs:1.15,0.5) {Intra-Class Distance};
    },
    extra description/.append code={
    \node[rotate=90, anchor=south, font=\scriptsize]
    at (rel axis cs:-0.14,0.5) {Inter/Intra Class Ratio};
},
]
\addplot[ybar, fill=cyan!20!white, draw=cyan!60!black,
 error bars/.cd, y dir=both, y explicit]
coordinates {
    (FedAvg,1.18) +- (0.52,0.52)
    (SupConFL,1.12) +- (0.16,0.16)
    (TripletFL,1.31) +- (0.13,0.13)
    (QuadrupletFL,1.29) +- (0.25,0.25)
    (MOON,1.00) +- (0.18,0.18)
    (FedQuad,1.28) +- (0.47,0.47)
};
\end{axis}

\begin{axis}[
    width=5cm, height=6cm,
    ymin=0, ymax=0.8,
    axis x line=none,
    axis y line*=right,
    yticklabel style={font=\scriptsize},
    symbolic x coords={FedAvg, SupConFL, TripletFL, QuadrupletFL, MOON, FedQuad},
    xtick=data,
    enlarge x limits=0.2,
    legend style={
        at={(0.96,0.94)}, anchor=north east,
        font=\tiny, draw=none, fill=white, fill opacity=1.0
    }
]
\addplot[thick, mark=*, mark options={fill=blue!60!black}, draw=blue!60!black]
coordinates {
    (FedAvg,0.74) (SupConFL,0.25) (TripletFL,0.2)
    (QuadrupletFL,0.21) (MOON,0.039) (FedQuad,0.029)
};
\addlegendimage{thick, mark=*, draw=none, mark options={fill=blue!60!black}}
\legend{\scriptsize Intra}
\end{axis}
\end{tikzpicture}
&
% ---------- (c) ----------
\begin{tikzpicture}[baseline]
\begin{axis}[
    title={\footnotesize $\alpha=0.3$},
    width=5cm, height=6cm,
    ymin=0, ymax=2.5,
    axis y line*=left,
    xlabel={\footnotesize Methods},
    xticklabel style={font=\scriptsize, rotate=45, anchor=east},
    xlabel style={at={(axis description cs:0.5,-0.18)}, anchor=north},
    symbolic x coords={FedAvg, SupConFL, TripletFL, QuadrupletFL, MOON, FedQuad},
    xtick=data,
    bar width=10pt,
    ymajorgrids=true,
    grid style={dashed,gray!40},
    enlarge x limits=0.2,
    extra description/.append code={
        \node[rotate=-90, anchor=south, font=\scriptsize]
        at (rel axis cs:1.15,0.5) {Intra-Class Distance};
    },
    extra description/.append code={
    \node[rotate=90, anchor=south, font=\scriptsize]
    at (rel axis cs:-0.14,0.5) {Inter/Intra Class Ratio};
},
]
\addplot[ybar, fill=cyan!20!white, draw=cyan!60!black,
 error bars/.cd, y dir=both, y explicit]
coordinates {
    (FedAvg,1.16) +- (0.21,0.21)
    (SupConFL,1.13) +- (0.18,0.18)
    (TripletFL,1.31) +- (0.34,0.34)
    (QuadrupletFL,1.29) +- (0.63,0.63)
    (MOON,1.21) +- (0.02,0.02)
    (FedQuad,1.28) +- (0.14,0.14)
};
\end{axis}

\begin{axis}[
    width=5cm, height=6cm,
    ymin=0, ymax=0.8,
    axis x line=none,
    axis y line*=right,
    yticklabel style={font=\scriptsize},
    symbolic x coords={FedAvg, SupConFL, TripletFL, QuadrupletFL, MOON, FedQuad},
    xtick=data,
    enlarge x limits=0.2,
     legend style={
        at={(0.96,0.94)}, anchor=north east,
        font=\tiny, draw=none, fill=white, fill opacity=1.0
    }
]
\addplot[thick, mark=*, mark options={fill=blue!60!black}, draw=blue!60!black]
coordinates {
    (FedAvg,0.75) (SupConFL,0.24) (TripletFL,0.19)
    (QuadrupletFL,0.21) (MOON,0.08) (FedQuad,0.031)
};
\addlegendimage{thick, mark=*, draw=none, mark options={fill=blue!60!black}}
\legend{\scriptsize Intra}
\end{axis}
\end{tikzpicture}

\end{tabular}
\caption{
Inter/Intra-class ratio (↑ better) across federated learning methods on CIFAR-100  
(200 clients). Error bars indicate standard deviation over runs.
}
\label{fig:ratioc100}
\end{figure}

%%%%%%%%%%%%%%%%%%%%%%%%%%%%%%%%%%%%%%%%%%%%%%%%
% Tiny-Imagenet
%%%%%%%%%%%%%%%%%%%%%%%%%%%%%%%%%%%%%%%%%%%%%%%%
\begin{figure}[htbp]
\raggedright
\setlength{\tabcolsep}{6.5pt} % adjust spacing between columns
\renewcommand{\arraystretch}{1.0}
\begin{tabular}{ccc}

% ---------- (a) ----------
\begin{tikzpicture}[baseline]
\begin{axis}[
    title={\footnotesize IID},
    width=5cm, height=6cm,
    ymin=0, ymax=2.5,
    axis y line* = left,
    %ylabel={\footnotesize Inter/Intra Class Ratio},
    %yticklabel style={font=\scriptsize},
    %every axis y label/.append style={xshift=-0.7ex},
    xlabel={\footnotesize Methods},
    xticklabel style={font=\scriptsize, rotate=45, anchor=east},
    xlabel style={at={(axis description cs:0.5,-0.18)}, anchor=north},
    symbolic x coords={FedAvg, SupConFL, TripletFL, QuadrupletFL, MOON, FedQuad},
    xtick=data,
    bar width=10pt,
    ymajorgrids=true,
    grid style={dashed,gray!40},
    enlarge x limits=0.2,
    % --- right-side title ---
    extra description/.append code={
        \node[rotate=-90, anchor=south, font=\scriptsize]
        at (rel axis cs:1.15,0.5) {Intra-Class Distance};
    },
    extra description/.append code={
    \node[rotate=90, anchor=south, font=\scriptsize]
    at (rel axis cs:-0.14,0.5) {Inter/Intra Class Ratio};
},
]
\addplot[ybar, fill=cyan!20!white, draw=cyan!60!black,
 error bars/.cd, y dir=both, y explicit]
coordinates {
    (FedAvg,1.16) +- (0.03,0.03)
    (SupConFL,1.15) +- (0.99,0.99)
    (TripletFL,1.27) +- (0.01,0.01)
    (QuadrupletFL,1.23) +- (0.12,0.12)
    (MOON,1.17) +- (0.01,0.01)
    (FedQuad,1.18) +- (0.13,0.13)
};
\end{axis}

% --- Line overlay ---
\begin{axis}[
    width=5cm, height=6cm,
    ymin=0, ymax=0.95,
    axis x line=none,
    axis y line*=right,
    yticklabel style={font=\scriptsize},
    symbolic x coords={FedAvg, SupConFL, TripletFL, QuadrupletFL, MOON, FedQuad},
    xtick=data,
    enlarge x limits=0.2,
    legend style={
        at={(0.96,0.94)}, anchor=north east,
        font=\tiny, draw=none, fill=white, fill opacity=1.0
    }
]
\addplot[thick, mark=*, mark options={fill=blue!60!black}, draw=blue!60!black]
coordinates {
    (FedAvg,0.89) (SupConFL,0.56) (TripletFL,0.29)
    (QuadrupletFL,0.35) (MOON,0.38) (FedQuad,0.40)
};
\addlegendimage{thick, mark=*, draw=none, mark options={fill=blue!60!black}}
\legend{\scriptsize Intra}
\end{axis}
\end{tikzpicture}
&
% ---------- (b) ----------
\begin{tikzpicture}[baseline]
\begin{axis}[
    title={\footnotesize $\alpha=0.5$},
    width=5cm, height=6cm,
    ymin=0, ymax=2.5,
    axis y line*=left,
    xlabel={\footnotesize Methods},
    xticklabel style={font=\scriptsize, rotate=45, anchor=east},
    xlabel style={at={(axis description cs:0.5,-0.18)}, anchor=north},
    symbolic x coords={FedAvg, SupConFL, TripletFL, QuadrupletFL, MOON, FedQuad},
    xtick=data,
    bar width=10pt,
    ymajorgrids=true,
    grid style={dashed,gray!40},
    enlarge x limits=0.2,
    extra description/.append code={
        \node[rotate=-90, anchor=south, font=\scriptsize]
        at (rel axis cs:1.15,0.5) {Intra-Class Distance};
    },
    extra description/.append code={
    \node[rotate=90, anchor=south, font=\scriptsize]
    at (rel axis cs:-0.14,0.5) {Inter/Intra Class Ratio};
},
]
\addplot[ybar, fill=cyan!20!white, draw=cyan!60!black,
 error bars/.cd, y dir=both, y explicit]
coordinates {
    (FedAvg,1.13) +- (0.87,0.87)
    (SupConFL,1.15) +- (0.36,0.36)
    (TripletFL,1.27) +- (0.31,0.31)
    (QuadrupletFL,1.07) +- (0.65,0.65)
    (MOON,1.18) +- (0.32,0.32)
    (FedQuad,1.17) +- (0.37,0.37)
};
\end{axis}

\begin{axis}[
    width=5cm, height=6cm,
    ymin=0, ymax=0.95,
    axis x line=none,
    axis y line*=right,
    yticklabel style={font=\scriptsize},
    symbolic x coords={FedAvg, SupConFL, TripletFL, QuadrupletFL, MOON, FedQuad},
    xtick=data,
    enlarge x limits=0.2,
    legend style={
        at={(0.96,0.94)}, anchor=north east,
        font=\tiny, draw=none, fill=white, fill opacity=1.0
    }
]
\addplot[thick, mark=*, mark options={fill=blue!60!black}, draw=blue!60!black]
coordinates {
    (FedAvg,0.90) (SupConFL,0.40) (TripletFL,0.35)
    (QuadrupletFL,0.39) (MOON,0.42) (FedQuad,0.37)
};
\addlegendimage{thick, mark=*, draw=none, mark options={fill=blue!60!black}}
\legend{\scriptsize Intra}
\end{axis}
\end{tikzpicture}
&
% ---------- (c) ----------
\begin{tikzpicture}[baseline]
\begin{axis}[
    title={\footnotesize $\alpha=0.3$},
    width=5cm, height=6cm,
    ymin=0, ymax=2.5,
    axis y line*=left,
    xlabel={\footnotesize Methods},
    xticklabel style={font=\scriptsize, rotate=45, anchor=east},
    xlabel style={at={(axis description cs:0.5,-0.18)}, anchor=north},
    symbolic x coords={FedAvg, SupConFL, TripletFL, QuadrupletFL, MOON, FedQuad},
    xtick=data,
    bar width=10pt,
    ymajorgrids=true,
    grid style={dashed,gray!40},
    enlarge x limits=0.2,
    extra description/.append code={
        \node[rotate=-90, anchor=south, font=\scriptsize]
        at (rel axis cs:1.15,0.5) {Intra-Class Distance};
    },
    extra description/.append code={
    \node[rotate=90, anchor=south, font=\scriptsize]
    at (rel axis cs:-0.14,0.5) {Inter/Intra Class Ratio};
},
]
\addplot[ybar, fill=cyan!20!white, draw=cyan!60!black,
 error bars/.cd, y dir=both, y explicit]
coordinates {
    (FedAvg,1.12) +- (0.83,0.83)
    (SupConFL,1.15) +- (0.08,0.08)
    (TripletFL,1.28) +- (0.02,0.02)
    (QuadrupletFL,1.23) +- (0.19,0.19)
    (MOON,1.18) +- (0.02,0.02)
    (FedQuad,1.16) +- (0.27,0.27)
};
\end{axis}

\begin{axis}[
    width=5cm, height=6cm,
    ymin=0, ymax=0.95,
    axis x line=none,
    axis y line*=right,
    yticklabel style={font=\scriptsize},
    symbolic x coords={FedAvg, SupConFL, TripletFL, QuadrupletFL, MOON, FedQuad},
    xtick=data,
    enlarge x limits=0.2,
     legend style={
        at={(0.96,0.94)}, anchor=north east,
        font=\tiny, draw=none, fill=white, fill opacity=1.0
    }
]
\addplot[thick, mark=*, mark options={fill=blue!60!black}, draw=blue!60!black]
coordinates {
    (FedAvg,0.91) (SupConFL,0.41) (TripletFL,0.36)
    (QuadrupletFL,0.39) (MOON,0.43) (FedQuad,0.37)
};
\addlegendimage{thick, mark=*, draw=none, mark options={fill=blue!60!black}}
\legend{\scriptsize Intra}
\end{axis}
\end{tikzpicture}

\end{tabular}
\caption{
Inter/Intra-class ratio (↑ better) across federated learning methods on Tiny-ImageNet (200 clients). Error bars indicate standard deviation over runs.
}
\label{fig:ratiotiny}
\end{figure}
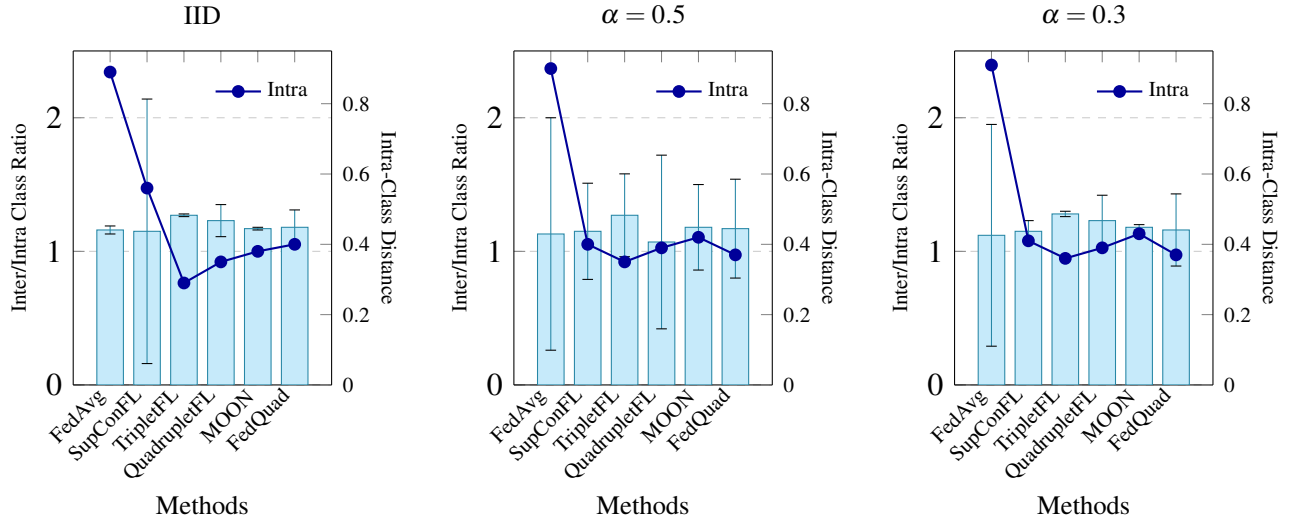

As an alternative, we evaluate these methods with stochastic client selection. The random client selection scenario highlights substantial variation in data distributions across training rounds. In this setting, a subset of clients is randomly selected at each communication round for model aggregation, while the remaining clients do not participate in updating their local models. Results show that increasing the total number of clients generally leads to a decline in accuracy, as the divergence between client data distributions becomes more pronounced, forcing the global model to aggregate highly inconsistent updates.
\begin{table}[htbp]
\small
\centering

\renewcommand{\arraystretch}{1.05}
\setlength{\tabcolsep}{4.5pt}

\begin{tabular}{c l cc cc cc}
\toprule
\multirow{2}{*}{\textbf{Participation $\%$}} &
\multirow{2}{*}{\textbf{Method}} &
\multicolumn{2}{c}{{CIFAR-10}} &
\multicolumn{2}{c}{{CIFAR-100}} &
\multicolumn{2}{c}{{Tiny-ImageNet}}\\
\cmidrule(lr){3-4}\cmidrule(lr){5-6}\cmidrule(lr){7-8}
 & & {200C} & {2000C} & {200C} & {2000C} & {200C} & {2000C} \\
\midrule

\multirow{6}{*}{$1\%$}
 & FedAvg                & 31.24{\tiny$\pm$3.64}  & 34.24{\tiny$\pm$3.18}  & 9.08{\tiny$\pm$1.08}  & 8.59{\tiny$\pm$0.73} & 5.11{\tiny$\pm0.23$}  & 3.97{\tiny$\pm$0.32}  \\
 & SupCon-FL             & 44.29{\tiny$\pm$0.27}  & 43.67{\tiny$\pm$0.44}  & 16.84{\tiny$\pm$0.2}  & 17.21{\tiny$\pm$0.28} & {11.56\tiny$\pm$0.31}  & 9.93{\tiny$\pm0.31$} \\
 & Triplet-FL            & 45.35{\tiny$\pm$0.6}  & 45.5{\tiny$\pm$0.53}  & 17.53{\tiny$\pm$0.3}  & 16.03{\tiny$\pm$0.34} & 12.47{\tiny$\pm0.51$}  & 10.57{\tiny$\pm0.30$} \\
 & Quadruplet-FL         & 52.96{\tiny$\pm$0.23}  & 50.61{\tiny$\pm$0.29}  & 16.01{\tiny$\pm$0.54}  & 15.92{\tiny$\pm$0.53} & 14.92{\tiny$\pm0.24$}  & 11.55{\tiny$\pm0.30$} \\
 & MOON                  & 28.99{\tiny$\pm$5.24}  & 15.59{\tiny$\pm$0.58}  & 5.16{\tiny$\pm$0.59}  & 1.45{\tiny$\pm$0.16} & 4.80{\tiny$\pm0.09$}  & 3.97{\tiny$\pm0.34$} \\
 & \textbf{FedQuad} & \textbf{53.03 {\tiny$\pm$0.81}}  & \textbf{50.39{\tiny$\pm$0.17}}  & \textbf{23.3{\tiny$\pm$0.36}}  & \textbf{18.95{\tiny$\pm$0.37}} & \textbf{15.00{\tiny$\pm0.16$}}  & \textbf{11.81{\tiny$\pm0.26$}} \\
\midrule

\multirow{6}{*}{$5\%$}
 & FedAvg                & 47.76{\tiny$\pm$2.1}  & 44.43{\tiny$\pm$1.01}  & 17.69{\tiny$\pm$0.39}  & 10.89{\tiny$\pm$0.69}  & 11.17{\tiny$\pm0.16$}  & 5.13{\tiny$\pm0.24$} \\
 & SupCon-FL             & 45.44{\tiny$\pm$0.34}  & 43.60{\tiny$\pm$0.55}  & 17.95{\tiny$\pm$0.25}  & 17.32{\tiny$\pm$0.2}  & 12.10{\tiny$\pm0.14$}  & 10.04{\tiny$\pm0.32$} \\
 & Triplet-FL            & 51.18{\tiny$\pm$0.21}  & 46.72{\tiny$\pm$0.27}  & 20.37{\tiny$\pm$0.27}  & 16.62{\tiny$\pm$0.29} & 14.01{\tiny$\pm0.4$}  & 11.16{\tiny$\pm0.33$} \\
 & Quadruplet-FL         & 57.62{\tiny$\pm$0.29}  & 51.75{\tiny$\pm$0.23}  & 18.81{\tiny$\pm$0.24}  & 16.97{\tiny$\pm$0.04} & 14.89{\tiny$\pm0.12$}  & 11.81{\tiny$\pm0.27$} \\
 & MOON                  & 33.94{\tiny$\pm$3.24}  & 15.09{\tiny$\pm$1.92}  & 8.33{\tiny$\pm$0.26}  & 2.13{\tiny$\pm$0.13} & 11.14{\tiny$\pm0.50$}  & 5.11{\tiny$\pm0.18$} \\
 & \textbf{FedQuad} & \textbf{58.21{\tiny$\pm$0.36}}  & \textbf{51.86{\tiny$\pm$0.39}}  & \textbf{26.47{\tiny$\pm$0.27}}  & \textbf{19.23{\tiny$\pm$0.6}} & \textbf{15.25{\tiny$\pm0.07$} } & \textbf{12.20{\tiny$\pm0.39$}} \\
\bottomrule
\end{tabular}
\caption{Method comparison on CIFAR-10, CIFAR-100, and Tiny-ImageNet with varying participation fraction and client scale.}
\label{tab:fraction_clients_comparison}
\end{table}

\begin{table}[ht]
\small
\centering
\renewcommand{\arraystretch}{1.2} % Optional: Increases row height for readability
\begin{tabular}{lllll}
\hline
\textbf{Method} &$\beta$ & $m_1$ & $m_2$ & {Accuracy ($\%$)}\\
\hline
    FedQuad & 0.5 &1.0 &  1.0& 71.51 \\
    FedQuad & 0.5 & 1.0 & 0.5& \textbf{72.68}\\
    FedQuad & 1.0 & 1.0 &  0.5& 70.4 \\
    FedQuad (without $\ell_{ce} $) & 0.5 & 1.0 & 0.5& 62.41\\
    FedQuad (without $\ell_{ce} $) & 0.5 & 2.0 & 0.5 & 60.99\\
    FedQuad (without $\ell_{ce} $) & 0.5 & 5.0 & 0.5 & 57.26\\
    FedQuad (without $\ell_{ce} $) & 0.5 & 1.0 & 1.0 & 56.42\\
\hline
\end{tabular}
\caption{Ablation study on the effect of loss hyperparameters ($\beta$, $m_1$, $m_2$) in the proposed quadruplet loss, evaluated on CIFAR-10 under an IID setting with 10 clients over 5 communication rounds.}
\label{tab:abl}
\end{table}
%%%%%%%%%%%%%%%%%%%%%%%%%%%
As demonstrated in Table \ref{tab:abl}, the inclusion of decision loss with optimal margin settings enhances the structural integrity of the feature space. Specifically, combining it with cross-entropy loss facilitates more effective quadruplet-based separation, achieving an accuracy score of 72.68 on CIFAR-10. This confirms that the proposed loss prevents representation collapse by enforcing wider inter-class margins.

However, increasing the participation fraction allows more clients to contribute in each round, as shown in Table \ref{tab:fraction_clients_comparison}, resulting in more stable aggregation and improved accuracy. This effect is particularly evident in quadruplet-based federated learning methods: Quadruplet FL achieves $57.62\%$ accuracy, whereas FedQuad attains $58.21\%$. These findings suggest that metric learning objectives, which explicitly optimize positive and negative pairwise relationships, are more robust under low client participation rates.

\section{Discussion}
\label{sec:6}
Our experimental results demonstrate that FedQuad consistently improves model generalisation in federated learning scenarios under varying degrees of data heterogeneity. In particular, FedQuad addresses one of the most critical challenges in FL, representational collapse which commonly arises under non-IID conditions. By explicitly optimising distances between positive and negative samples, our method increases inter-class variance, leading to more stable and discriminative global representations without requiring raw data exchange among clients.

While contrastive learning has shown strong performance in centralised settings, its effectiveness in federated learning is frequently limited, particularly under class-imbalanced distributions (e.g., Dirichlet $\alpha=0.3$). This limitation can be attributed to the inherent constraints of contrastive objectives, which may produce misaligned latent representations across clients. In contrast, our stochastic alignment strategy introduces greater flexibility into representation learning, enabling more effective adaptation across diverse client distributions and resulting in consistently improved performance, especially under severe non-IID conditions.

Moreover, traditional contrastive and triplet-based losses are prone to degradation under high class imbalance, often leading to noisy or overlapping feature embeddings. FedQuad mitigates this issue by preserving the structure of local representations while aligning them effectively within a global feature space. Notably, this is achieved without explicitly modelling distances between negative pairs or requiring any data sharing, thereby maintaining strict privacy constraints while ensuring globally coherent representations.

A key insight from our study is that achieving a balance between local discriminability and global consistency in federated learning requires careful loss function design. The proposed stochastic quadruplet formulation extends beyond conventional contrastive and triplet losses by introducing stronger negative constraints and finer-grained control over pairwise relationships. As a result, the learned embeddings are significantly more robust to client-level distributional shifts, which is particularly beneficial in low-resource or highly imbalanced settings where standard FL methods tend to fail.

Despite its strong performance, FedQuad has several limitations. Its applicability is restricted in scenarios where clients possess only a small number of classes (e.g., binary or few-class classification), as effective negative sample mining becomes infeasible. In particular, the proposed loss function requires samples from at least three distinct classes to construct valid quadruplets. Therefore, FedQuad does not apply to datasets with only one or two classes, where the required anchor–positive–multiple negative relationships cannot be properly formed. Furthermore, as shown in Table \ref{tab:tiny3}, increased diversity under high heterogeneity can lead to weaker representations and degraded performance, suggesting that a fixed architecture and preprocessing pipeline may not be optimal for highly diverse datasets such as TinyImageNet.

Finally, our approach focuses on simultaneously contrasting a positive sample against multiple negatives, enabling the model to push embeddings away from multiple directions while preserving intra-class similarity, an important factor in improving representation robustness. While experiments on CIFAR-10, CIFAR-100 and Tiny-ImageNet validate the effectiveness of FedQuad, extending evaluation to domain-specific and resource-constrained datasets (e.g., medical imaging or user behaviour data) would further demonstrate its practical applicability and generalizability in real-world federated learning scenarios.

\section{Conclusion}
In this work, we evaluate FedQuad, a metric learning-based federated learning framework designed to directly address representational collapse caused by data heterogeneity across clients. By leveraging a stochastic quadruplet loss, FedQuad promotes reduced intra-class variance and increased inter-class variance within local feature spaces, thereby enhancing the quality of global representations without requiring access to raw client data. In addition, we provide a detailed analysis of metric learning in federated settings, particularly under conditions of data imbalance and limited data availability with random client selection.

Extensive experiments on multiple benchmark datasets under diverse non-IID scenarios demonstrate that FedQuad effectively outperforms existing baselines, especially in the presence of severe class imbalance and a large number of clients. These findings highlight the effectiveness of metric learning for local representation alignment and emphasise the importance of structured embedding objectives in mitigating statistical heterogeneity.

Future research will focus on two key areas: first, the adaptation of FedQuad for unsupervised and semi-supervised settings to broaden its applicability; and second, the optimization of hard negative mining, which is essential for improving distance-based representation separation in highly heterogeneous feature spaces.\\

%\textbf{Acknowledgement}

\end{document}